\documentclass[11pt]{article}

\PassOptionsToPackage{table}{xcolor}
\usepackage[preprint]{acl}

\usepackage{times}
\usepackage{latexsym}

\usepackage[T1]{fontenc}

\usepackage[utf8]{inputenc}

\usepackage{microtype}

\usepackage{amsmath}
\usepackage{amssymb}
\usepackage{graphicx}
\usepackage{booktabs}
\usepackage{multirow}
\definecolor{gainbg}{HTML}{D9EAD3}
\definecolor{oursbg}{HTML}{D6E9F8}
\usepackage{placeins}
\usepackage{float}
\usepackage{tcolorbox}
\tcbuselibrary{breakable}

\title{Beyond Reward Engineering: A Data Recipe for Long-Context Reinforcement Learning}

\author{
Xiaoyue Xu$^{1}$,
Sikui Zhang$^{1}$,
Xiaorong Wang$^{1}$,
{Xu Han}$^{2}$\footnotemark[1],
Chaojun Xiao$^{2}$\thanks{Corresponding authors}\\
$^{1}$OpenBMB \quad $^{2}$Tsinghua University \\
\texttt{xiaoyue.xu.me@gmail.com}, \texttt{\{han-xu,xcj\}@tsinghua.edu.cn}
}

\begin{document}
\maketitle
\begin{abstract}
Long-context reasoning is an essential capability for large language models, particularly when they are deployed as autonomous agents that must reason over lengthy trajectories.
Reinforcement learning (RL) has recently emerged as a dominant paradigm for improving this ability, yet existing work largely focuses on reward engineering while diverse training data remains scarce.
We revisit this problem from a data-centric perspective and show that a simple yet effective data recipe alone, paired with a minimal outcome-based GRPO setup, suffices to substantially improve long-context reasoning.
Our recipe targets three complementary task families -- \textbf{retrieval}, \textbf{multi-evidence synthesis}, and \textbf{reasoning} -- for which we construct and curate eight datasets totaling $\sim$14K examples.
Experiments on three models (Qwen3-4B/8B/30B-A3B) yield average gains of $+7.2$/$+3.2$/$+6.4$ points across seven long-context benchmarks, surpassing prior RL training sets.
We further demonstrate that these gains transfer to agentic tasks, where continuing RL training on an agent-tuned model with our data recipe improves GAIA by $+4.8$ and BrowseComp by $+7.0$ points.
We release our datasets to facilitate future research.\footnote{\url{https://huggingface.co/datasets/openbmb/UltraData-RL-2609} (Long-Context Subset)}

\end{abstract}

\section{Introduction}
\label{sec:intro}

\begin{figure}[t!]
\centering
\includegraphics[width=0.8\columnwidth]{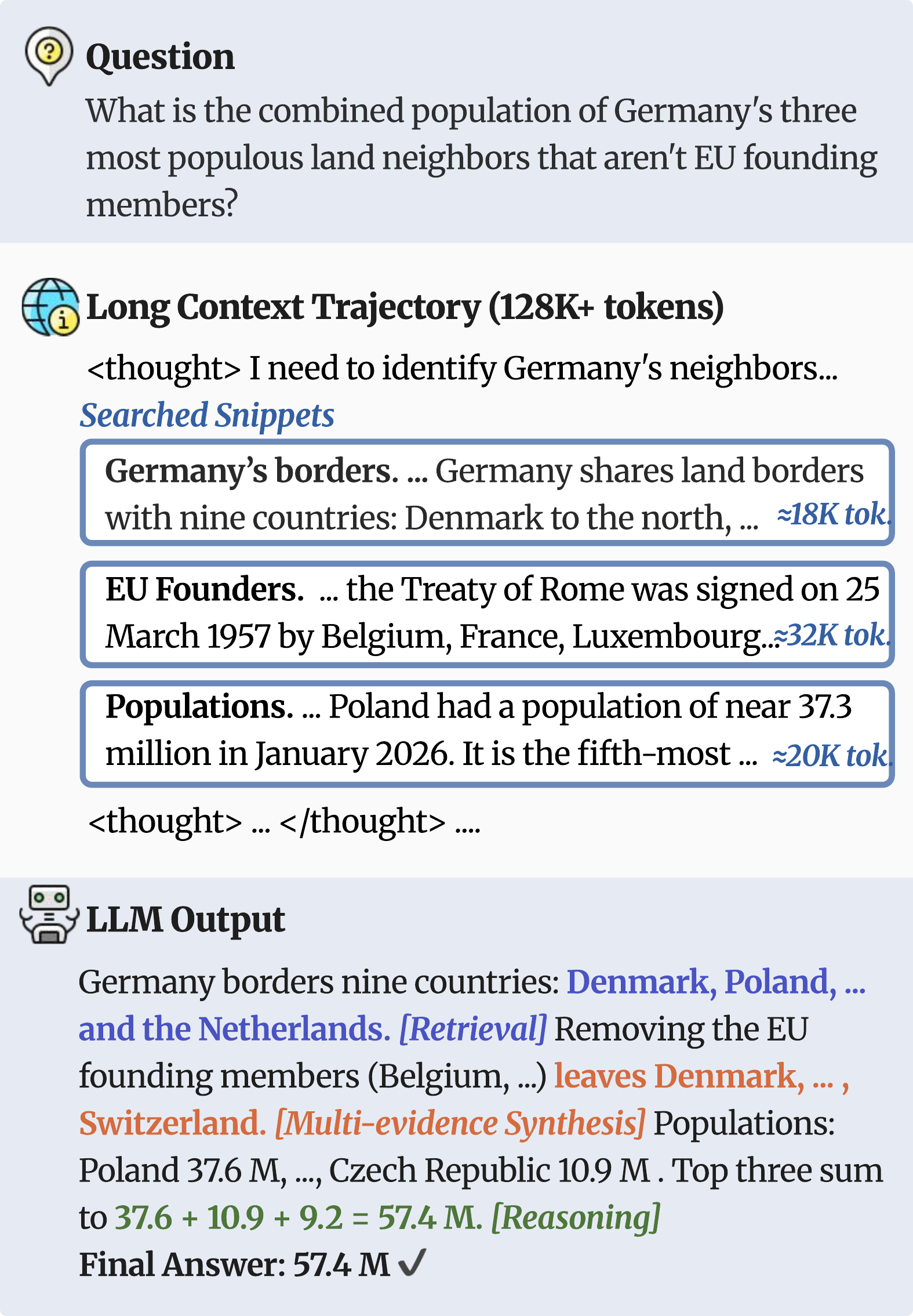}
\caption{A long-context reasoning example requiring three abilities that define our training mixture (\S\ref{sec:data}): \textbf{Retrieval}, \textbf{Multi-evidence Synthesis}, and \textbf{Reasoning}. }
\label{fig:teaser}
\end{figure}

Large language models (LLMs) have been increasingly deployed as autonomous agents, successfully tackling complex real-world tasks such as web searching~\citep{mialon2023gaia,wei2025browsecomp}, software engineering~\citep{yang2024sweagent}, and GUI automation~\citep{qin2025uitars}. These applications expose LLMs to extensive web pages, large-scale code repositories, and massive action histories, where task-relevant evidence is often scattered across lengthy inputs. LLMs must identify, integrate, and reason over this information to produce correct decisions. Long-context reasoning has therefore become an essential capability for modern LLMs.

Previous work has proven the effectiveness of reinforcement learning (RL) in enhancing the reasoning ability of LLMs~\citep{openai2024o1,deepseekai2025r1}, yet most efforts focus on short-context tasks. Extending such gains to long-context reasoning remains underexplored. Recent studies indicate that the outcome-only reward of standard Reinforcement Learning with Verifiable Rewards (RLVR) provides too sparse a signal to guide the model in locating evidence over long inputs, leading to plateaued contextual recall and shortcut reasoning that bypass grounding~\citep{chen2026longrlvr,guan2026eapo}.
In response, most existing efforts focus on reward design of long-context RL~\citep{ping2026longr,chen2026longrlvr,peng2026longpas,guan2026eapo}, supplementing the outcome reward with auxiliary signals that explicitly score evidence grounding or intermediate reasoning. Meanwhile, high-quality training data remains scarce: while several synthetic data pipelines have been proposed~\citep{shen2025qwenlongl15,wang2026loongrl,xiao2026decomposition}, the resulting datasets are either narrow in task coverage or kept closed-source.

In this paper, we revisit long-context RL from a data-centric perspective and show that a diverse data recipe alone is sufficient to substantially improve long-context reasoning, without special reward engineering. We hypothesize that long-context reasoning decomposes into a small set of complementary core abilities that need to be exercised jointly during training, a view echoed by prior empirical analyses~\citep{li2024alr2,xiao2026decomposition}. For example, given the web pages gathered to answer a GAIA-like~\citep{mialon2023gaia} question such as \emph{``What is the combined population of Germany's three most populous land neighbors that aren't EU founding members?''} (Figure~\ref{fig:teaser}), the model must (1) locate candidate countries together with their populations, (2) integrate the border-relationship and EU-founding-member information to filter out the founding members, and (3) perform math reasoning over the three most populous remaining countries to compute the final sum.

Guided by this hypothesis, our data recipe spans three corresponding task categories: (i)~\textbf{Retrieval}, where the model must locate relevant evidence from a long input, especially when the evidence is semantically indirect, or the information is scattered among abundant distractors; (ii)~\textbf{Multi-evidence Synthesis}, where the model must integrate evidence from different parts of the input to derive an answer that no single piece can provide; and (iii)~\textbf{Reasoning}, where the model must perform complex multi-step problem solving (e.g., math) over a long input. We accordingly collect and construct training data covering these three task categories, and conduct thorough RL experiments to evaluate the effectiveness of our data recipe on long-context benchmarks as well as its transfer to agentic tasks.

We train on three models, Qwen3-4B-Thinking-2507, Qwen3-8B, and Qwen3-30B-A3B-Thinking-2507 ~\citep{qwen3}, and evaluate across seven long-context benchmarks. Without any special reward engineering or staged RL training, our recipe delivers consistent gains, achieving average improvements of $+7.2$, $+3.2$, and $+6.4$ points on the three models respectively. Under the same RL setup, it outperforms two prior long-context RL training sets, DocQA-RL-1.6K~\citep{wan2025qwenlongl1} and KeyChain~\citep{wang2026loongrl}.
We further show that these long-context gains transfer to agentic tasks, as continuing long-context RL on an agent-tuned model improves GAIA by $4.8$ and BrowseComp by $7.0$ points, suggesting that strengthening core long-context abilities effectively benefits long-horizon agentic capability.


\section{Related Works}

\paragraph{Long-context Reasoning via Post-training.}
A growing line of work improves long-context reasoning through post-training, where a central challenge is curating diverse and high-quality long-context supervision.
On the SFT side, LongAlign~\citep{bai2024longalign} builds a long instruction-following dataset paired with a balanced training strategy, \citet{li2024selfimprovelong} allows the model to generate its own SFT examples for self-improvement, and ACC~\citep{su2026acc} compiles long agent trajectories into long-context training data.
Reinforcement learning has more recently emerged as an effective paradigm for optimizing reasoning behavior under long context, which we discuss next.

\paragraph{Reinforcement Learning in Long-context Scenarios.}
Existing efforts in long-context RL can be broadly divided into two lines: \emph{algorithm-centric} and \emph{data-centric}.
The algorithm-centric line improves the RL training procedure.
On the reward side, recent works supplement the outcome-only reward of RLVR with auxiliary signals that explicitly score evidence grounding~\citep{chen2026longrlvr, guan2026eapo, whitecross2026recallm} or intermediate-step quality~\citep{peng2026longpas}.
On the optimization side, it includes modified policy optimization~\citep{shen2025qwenlongl15} and selective weight updates~\citep{ping2026longact}.
The data-centric line constructs high-quality long-context RL training sets, such as DocQA-RL~\citep{wan2025qwenlongl1}, which contains $\sim$1.6K document QA problems, LoongRL~\citep{wang2026loongrl}, which synthesizes examples through a KeyChain UUID-driven pipeline, and \citet{xiao2026decomposition}, which builds a dataset via atomic-skill decomposition. Yet their training data is either limited in scale and diversity or confined to a single synthetic task format.

Our work focuses on the data-centric line.
Different from these prior efforts, we contribute a more unified \emph{recipe} that exercises three complementary core abilities of long-context reasoning, yielding consistent gains across various long-context benchmarks.


\begin{table*}[!t]
\centering
\small
\setlength{\tabcolsep}{6pt}
\renewcommand{\arraystretch}{1.15}
\begin{tabular}{@{}llp{8.5cm}rc@{}}
\toprule
\textbf{Category} & \textbf{Dataset} & \textbf{Description} & \textbf{Size} & \textbf{Length} \\
\midrule
\multirow{2}{*}{Retrieval}
  & FuzzyNeedle  & IS-A paraphrased needles, parent-class queries     & 1{,}500 & 0--32K \\
  & MultiNeedle  & Shuffled near-duplicate multiple needle retrieval           & 1{,}500 & 0--32K \\
\midrule
\multirow{5}{*}{Multi-Evidence}
  & CrossEntity & Cross-entity aggregation over derived attributes   & 3{,}521 & 0--64K \\
  & WebSearch    & Bridging chains with sibling-entity distractors    & 684     & 0--64K \\
  & MultiQuery   & Embedded queries answered in order                 & 226     & 0--32K \\
  & KeyChain     & UUID pointer chains hiding the question            & 654     & 0--32K \\
  & LongDocQA    & Extended from HotpotQA / MuSiQue / Qasper          & 3{,}422 & 0--64K \\
\midrule
Reasoning
  & LongMath     & Math problems with scattered variables             & 2{,}562 & 0--64K \\
\midrule
\end{tabular}
\caption{Composition of our long-context training mixture across the three ability categories. ``Length'' denotes the estimated input length range in tokens.}
\label{tab:data_recipe}
\end{table*}

\section{Methodology}

Our methodology has two parts: a long-context training data mixture that jointly stresses three core abilities of long-context reasoning (\S\ref{sec:data}), and a minimal RL setup with an outcome-based reward (\S\ref{sec:rl}).

\subsection{Long-Context Training Data}
\label{sec:data}

Our training mixture is organized around the three task categories introduced in \S\ref{sec:intro}---\textbf{Retrieval}, \textbf{Multi-evidence Synthesis}, and \textbf{Reasoning}---each isolating a distinct ability that is essential for long-context reasoning. For each category, we analyze the main difficulties models face and introduce the corresponding dataset(s) we construct to address them. We then conclude with the overall data recipe and describe our data curation process.

\subsubsection{Retrieval}
Vanilla needle-in-a-haystack~\citep{kamradt2023niah} is largely saturated
on recent long-context models. However, information retrieval in current long-context LMs still relies heavily on lexical pattern
matching~\citep{modarressi2025nolima, xu2024lifelongicl}, and degrades
sharply once the relevant information becomes abundant and scattered
across the context (e.g., as the number of needles
grows)~\citep{hsieh2024ruler, vodrahalli2024mrcr, li2024needlebench}. We
therefore construct two lightweight datasets that target these
two shortcomings.

\paragraph{FuzzyNeedle.}
FuzzyNeedle removes lexical shortcuts by paraphrasing the needle through a Wikidata \emph{IS-A} relation so that the query and the evidence share no keywords, in a similar spirit to NoLiMa~\citep{modarressi2025nolima}.
We first sample a target entity (e.g., \texttt{fish}) and an instance of it (e.g., \texttt{salmon}), and attach this instance to a fictional statement (e.g., \emph{``Alice loves the texture of salmon''}). 
We then additionally sample sibling entities (e.g., \texttt{vegetable}) and draw several instances to form analogous distractors for each of them.
All statements are scattered throughout an unrelated long passage, and the model is queried at the target entity level (e.g., \emph{``Who enjoys eating fish?''}).
Answering correctly requires resolving the is-instance-of relation
rather than relying on keyword matching, as well as rejecting the distractor statements.

\paragraph{MultiNeedle.}
MultiNeedle requires the model to distinguish among many near-duplicate needles and
retrieve a target one by its order of occurrence, following
MRCR~\citep{vodrahalli2024mrcr}. We prepare a set of diverse topics, and prompt LLMs to draft multiple user--model conversations on each topic. Conversations from different topics are then shuffled and joined into a single long context, and the model is asked to retrieve the \emph{$K$-th} conversation on a specified topic.
Models thus need to exhaustively locate every needle, keep a count of their order and discriminate among similar candidates.

\subsubsection{Multi-evidence Synthesis}
Multi-evidence synthesis is a common pattern in real-world long-context tasks~\citep{yang2018hotpotqa, tang2024multihoprag}, which requires integrating information scattered across different parts of the input into a coherent answer.
Long-context LMs exhibit several shortcomings here:
(i)~\emph{surface-form shortcuts}, utilizing only verbatim spans rather than synthesizing the evidence; 
(ii)~\emph{incomplete coverage}, where the model skips parts of the long input and
misses information needed for the answer~\citep{gupta2025novelhopqa}; 
and (iii)~\emph{distractor confusion}, where topically related
documents are mistaken for gold evidence~\citep{vodrahalli2024mrcr}. 
For this kind of task, we construct new synthetic datasets targeting these shortcomings, and adapt long-context QA sources from prior work.

\paragraph{CrossEntity.}
This dataset targets both surface-form shortcuts and incomplete coverage by producing questions whose answer is an aggregation over \emph{derived} attributes of multiple entities, which are never stated directly in the documents. Given a group of peer entities from the same Wikidata category (e.g., NBA players), we prompt LLMs to identify both \emph{direct attributes} mentioned in the entity pages (e.g., \texttt{birth\_year}) and \emph{derived attributes} obtainable only by synthesizing over them (e.g., $\texttt{winning\_age}=\texttt{winning\_year}-\texttt{birth\_year}$). We then generate cross-entity comparison questions about the derived attributes (e.g., \emph{``Which of these players was the youngest at the time of the first championship?''}). 
Models are challenged to first integrate evidence within each individual document, and then perform a global aggregation across all entities without any omission.

\paragraph{WebSearch.}
Targeting distractor confusion, WebSearch draws inspiration from search-agent data construction~\citep{liu2025webexplorer, lu2025deepdive}, where referencing a target entity through a sequence of bridging relations naturally yields a multi-hop question. We first collect relation triples whose entities are linked into a single chain, and substitute most entities with letter placeholders to generate a question, e.g., \emph{``A is the director of B; the lead actor of B is C; \dots; who is A?''}. Web pages of the gold triples form the supporting documents. We then prompt LLMs to brainstorm similar entities satisfying the same relation (e.g., \emph{``E is the director of F''}) and collect their pages as distractors. All documents are interleaved in a random order. The models must trace the chain hop by hop, and distinguish the documents that exactly match each bridging relation from topically similar distractors.

\paragraph{MultiQuery.}
To stress incomplete coverage, we introduce a straightforward task in which missing any piece of evidence directly causes an incorrect response. We sample documents from pre-training corpus and prompt LLMs to generate QA pairs. Queries are then randomly embedded in the concatenated documents and models are asked to answer all questions in the order they appear. In this task, each question constitutes a required piece of evidence, and failing to gather any of them makes the final response incomplete.

\paragraph{KeyChain.}
We also construct a KeyChain dataset following the pipeline of LoongRL~\citep{wang2026loongrl}. We generate QA pairs from the same corpus as MultiQuery, hide each question behind a chain of UUID pointers, and scatter the chains throughout the assembled multi-document context. The model must trace the chain from a starting UUID and recover the actual question before answering.

\paragraph{LongDocQA.}
We include the training sets of HotpotQA~\citep{yang2018hotpotqa}, MuSiQue~\citep{trivedi2022musique}, and Qasper~\citep{dasigi2021qasper}, and extend their length to further improve data diversity. For HotpotQA, we replace the truncated supporting paragraphs with the corresponding full Wikipedia articles; for MuSiQue and Qasper, we keep the original input intact and pad it with irrelevant texts.

\subsubsection{Reasoning}
Previous works observe that long inputs alone degrade a model's reasoning performance~\citep{levy2024sametask, du2025contextlength}, which motivate us to train LLMs on long-form reasoning tasks so as to preserve deep thinking ability when inputs get lengthy. 
Training data of such reasoning-intensive problem is naturally scarce in long-context setting, while comparably abundant in short-context math and code
domains~\citep{hendrycks2021math, deepseekai2025r1}. 
We then address this gap by adapting difficult short-context math problems into the long-context format.

\paragraph{LongMath.}
We sample difficult problems from two challenging math datasets, DeepMath-103K and MATH-Hard~\citep{he2025deepmath, hendrycks2021math}.
Following the spirit of LongReason~\citep{longreason}, we prompt an LLM to rewrite each question into a narrative scenario (e.g., science fiction, finance, or engineering), where the original variables and numerical conditions are distributed across at least five story fragments. A second LLM verifies the rewrites and filters out the unsolvable ones. Finally, the fragments are scattered throughout an irrelevant long document, and the model must locate all of them and perform complex mathematical derivation on top of this lengthy input.

\subsubsection{Data Curation and Final Recipe}
Across all sources, we apply a two-stage post-processing pipeline. We first discard samples exceeding $64$K tokens, in order to keep rollout cost tractable while preserving sufficient long-context difficulty. We then rollout each remaining question four times with Qwen3-30B-A3B-Thinking-2507~\citep{qwen3} and retain only those that are neither all-correct nor all-wrong, isolating problems of moderate difficulty that yield informative learning signals for RL. This yields a total of $14{,}069$ examples, as detailed in Table~\ref{tab:data_recipe}.

\subsection{Long-Context Reinforcement Learning}
\label{sec:rl}

\paragraph{GRPO objective.}
We train with Group Relative Policy Optimization (GRPO)~\citep{shao2024deepseekmath}. For each prompt $x$, GRPO samples a group of $G$ rollouts $\{y_i\}_{i=1}^{G}$ from $\pi_{\theta_{\mathrm{old}}}$, scores them with rewards $\{r_i\}_{i=1}^{G}$, and assigns each rollout the group-relative advantage
\begin{equation}
\label{eq:adv}
A_i = \frac{r_i - \operatorname{mean}(\{r_j\}_{j=1}^{G})}{\operatorname{std}(\{r_j\}_{j=1}^{G})}.
\end{equation}
GRPO then forms the per-token surrogate loss
\begin{equation}
\label{eq:loss}
\begin{split}
\mathcal{L}_{i,t}(\theta) = \min\bigl(&\rho_{i,t}\, A_i,\\
&\operatorname{clip}(\rho_{i,t},1{-}\varepsilon,1{+}\varepsilon)\, A_i\bigr),
\end{split}
\end{equation}
where
\begin{equation}
\label{eq:ratio}
\rho_{i,t} = \frac{\pi_\theta(y_{i,t}\mid x,y_{i,<t})}{\pi_{\theta_{\mathrm{old}}}(y_{i,t}\mid x,y_{i,<t})},
\end{equation}
and maximizes the objective
\begin{equation}
\label{eq:grpo}
\mathcal{J}_{\mathrm{GRPO}}(\theta) = \mathbb{E}_{x,\{y_i\}}\!\biggl[\frac{1}{G}\sum_{i=1}^{G}\frac{1}{|y_i|}\sum_{t=1}^{|y_i|}\mathcal{L}_{i,t}(\theta)\biggr].
\end{equation}
Following recent practice in reasoning RL~\citep{yu2025dapo}, we omit the KL regularization term.

\paragraph{Multi-dataset stabilization.}
Our training mixture combines $8$ heterogeneous datasets whose reward distributions differ in both scale and variance. To stabilize training, we follow QwenLong-L1.5~\citep{shen2025qwenlongl15} and apply (i) \emph{task-balanced sampling}, which fixes the per-dataset sample ratio in each batch, and (ii) \emph{task-level advantage normalization}, which rescales each rollout's advantage by the per-dataset reward standard deviation so that low-variance datasets are not washed out by high-variance ones.

\paragraph{Reward design.}
We use a simple reward design. For most datasets, we use a word-level recall reward $r=|\hat{y}\cap y^\star|/|y^\star|$ between the model's final answer $\hat{y}$ and the label $y^\star$. For CrossEntity and LongMath, we use DeepSeek-V3.2 as a judge, as recall reward can be hacked by enumerating candidates for CrossEntity and fails to credit equivalent forms of LongMath's numerical answers.


\begin{table*}[!t]
\centering
\small
\setlength{\tabcolsep}{4pt}
\renewcommand{\arraystretch}{1.1}
\resizebox{\textwidth}{!}{%
\begin{tabular}{l c cc ccc cc}
\toprule
\multirow{2}{*}{\textbf{Model}} & \multirow{2}{*}{\textbf{Avg.}} & \multicolumn{2}{c}{\textbf{Multi-hop QA}} & \multicolumn{3}{c}{\textbf{Holistic LC Reasoning}} & \multicolumn{2}{c}{\textbf{Synthetic LC Reasoning}} \\
\cmidrule(lr){3-4} \cmidrule(lr){5-7} \cmidrule(lr){8-9}
 & & LBv1-QA & FRAMES & LBv2 & AA-LCR & DocFinQA & LongReason & GraphWalks \\
\midrule
Qwen3-4B-Thinking-2507  & 53.42          & 62.60          & 77.55          & 40.60          & 35.25          & 39.80          & 69.50          & 48.67 \\
\quad + DocQA-RL-1.6K    & 55.08          & 63.80          & 78.28          & 42.25          & 32.50          & 48.05          & 71.28          & 49.42 \\
\quad + KeyChain-15K     & 57.97          & 66.80          & 82.40          & 43.39          & 41.00          & 47.29          & 76.20          & 48.70 \\
\rowcolor{oursbg}
\quad + \textbf{Ours}    & \textbf{60.63} & \textbf{68.30} & \textbf{82.52} & \textbf{47.60} & \textbf{45.75} & \textbf{48.37} & \textbf{82.12} & \textbf{49.74} \\
\midrule
Qwen3-8B-128K           & 49.93          & 62.90          & 74.76          & 33.75          & 28.75          & 38.83          & 68.26          & \textbf{42.27} \\
\quad + DocQA-RL-1.6K    & 49.08          & 63.70          & 71.36          & 34.84          & 28.50          & 37.96          & 67.88          & 39.29 \\
\quad + KeyChain-15K     & 52.08          & 66.50          & \textbf{80.22} & 34.14          & \textbf{34.50} & 35.79          & 72.04          & 41.36 \\
\rowcolor{oursbg}
\quad + \textbf{Ours}    & \textbf{53.10} & \textbf{68.80} & 77.67          & \textbf{36.28} & \textbf{34.50} & \textbf{39.59} & \textbf{73.93} & 40.96 \\
\midrule
Qwen3-30B-A3B-Thinking-2507 & 61.82          & 66.50          & 79.37          & 49.40          & 48.75          & 41.11          & 81.61          & 66.00 \\
\quad + DocQA-RL-1.6K    & 64.02          & 68.90          & 81.55          & 53.18          & 43.30          & 51.41          & 80.86          & \textbf{68.91} \\
\quad + KeyChain-15K     & 64.57          & 68.40          & 85.07          & 52.69          & 45.25          & 50.00          & 84.76          & 65.82 \\
\rowcolor{oursbg}
\quad + \textbf{Ours}    & \textbf{68.23} & \textbf{77.30} & \textbf{85.80} & \textbf{54.21} & \textbf{56.00} & \textbf{51.63} & \textbf{84.89} & 67.75 \\
\bottomrule
\end{tabular}%
}
\caption{Main results on seven long-context benchmarks. \textbf{Bold} denotes the best score per column within each base model. Our data recipe delivers consistent improvements across all benchmarks and base models, surpassing the two baselines.}
\label{tab:main_results}
\end{table*}

\section{Experiments}
\label{sec:experiments}

\subsection{Setup}

\paragraph{Training Details.}
We conduct reinforcement learning on three base models, Qwen3-4B-Thinking-2507~\citep{qwen3-4b-thinking-2507}, Qwen3-8B~\citep{qwen3-8b-card}, and Qwen3-30B-A3B-Thinking-2507~\citep{qwen3-30b-a3b-thinking-2507}, which are commonly adopted by recent long-context RL works \citep{ping2026longr,ping2026longact,peng2026longpas}. 
For Qwen3-8B, we extend the context window from $32{,}768$ to $131{,}072$ tokens via YaRN~\citep{peng2023yarn}, following the official recommendation of ~\citep{qwen3-8b-card}.
We perform all RL training using the Miles framework~\citep{miles2025}. 
For each prompt we sample $G=8$ rollouts with temperature $1.0$ and top-$p=0.95$, capping the input length at $64$K tokens and the generation length at $16$K tokens.
The policy is updated with the Adam optimizer at a constant learning rate ($1\text{e-}6$ by default; see Appendix~\ref{sec:hyperparams} for per-model settings), using a global batch size of $128$ prompts and a PPO batch size of $1{,}024$ (i.e., on-policy RL).

\paragraph{Benchmarks.}
We evaluate on a comprehensive suite of seven long-context benchmarks, organized into three groups according to the primary abilities they probe:
(i)~Multi-hop QA on real-world documents, covering LongBench~v1 QA \citep{bai2024longbench}, from which we follow \citet{wan2025qwenlongl1} and \citet{wang2026loongrl} and use the five document-QA subsets (HotpotQA, 2WikiMultiHopQA, MuSiQue, NarrativeQA, and Qasper), and FRAMES \citep{gemini2024frames};
(ii)~Holistic Long-Context Reasoning, covering LongBench~v2 \citep{bai2025longbenchv2}, AA-LCR \citep{aalcr2024}, and DocFinQA \citep{docfinqa};
(iii)~Synthetic Long-Context Reasoning, covering LongReason \citep{longreason} and GraphWalks \citep{graphwalks}.
We use a unified decoding configuration with temperature $0.6$, top-$p=0.95$, a maximum input length of $130$K tokens, and a maximum output length of $60$K tokens.
For LongBench~v1 QA and FRAMES, we report the maximum of exact match and LLM-judged accuracy as the final score. For LongBench~v2 and AA-LCR, we report Avg@4 to reduce evaluation variance.

\paragraph{Baselines.}
We compare our data recipe against two existing long-context RL training datasets:
(1) DocQA-RL-1.6K~\citep{wan2025qwenlongl1}, the RL training set of QwenLong-L1, consisting of 1{,}591 long-document QA problems across three reasoning types ;
(2) KeyChain-15K~\citep{wang2026loongrl}, the open-sourced LoongRL training data synthesized by the KeyChain UUID-driven pipeline, comprising $\sim$15K multi-hop QA examples built on HotpotQA, MuSiQue, and 2WikiMQA and uniformly preprocessed to a $16$K-token context. 


\subsection{Main Results}

\paragraph{Improvements across benchmarks and models.}
Table~\ref{tab:main_results} summarizes our main results across three models.
We highlight two observations. (i) Training with our data recipe yields consistent improvements : all seven benchmarks improve on Qwen3-4B-Thinking-2507 and Qwen3-30B-A3B-Thinking-2507, and six out of seven improve on Qwen3-8B-128K. In particular, significant gains are observed on LBv2, AA-LCR, and DocFinQA, three challenging benchmarks that evaluate holistic long-context reasoning ($+7.0$, $+10.5$, and $+8.6$ respectively for Qwen3-4B-Thinking-2507; $+4.8$, $+7.3$, and $+10.5$ for Qwen3-30B-A3B-Thinking-2507). (ii) Models of different sizes all benefit from our recipe, with average score gains of $+7.2$, $+3.2$, and $+6.4$ for Qwen3-4B-Thinking-2507, Qwen3-8B-128K, and Qwen3-30B-A3B-Thinking-2507 respectively. The consistent gains across model sizes and variants indicate the broad applicability of our training recipe.

\paragraph{Comparison with prior long-context RL data.}
Table~\ref{tab:main_results} also compares our recipe against the two baseline datasets on all three models. Our data recipe achieves the highest average score on every model, outperforming the stronger KeyChain-15K baseline by $+2.66$, $+1.02$, and $+3.66$ on the 4B, 8B, and 30B-A3B models respectively. Moreover, our data recipe yields the largest gains on most benchmarks across three models, especially on the reasoning-heavy benchmarks (i.e., LBv2, AA-LCR, and LongReason).

\begin{figure}[t]
\centering
\includegraphics[width=\columnwidth]{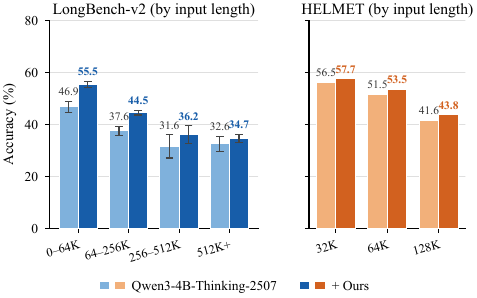}
\caption{Training on our max-64K-token data recipe generalizes to substantially longer contexts. \textbf{Left}: LongBench-v2 accuracy stratified by input length. \textbf{Right}: HELMET accuracy at 32K, 64K, and 128K input lengths. Error bars on LongBench-v2 show the standard deviation over four sampling seeds.}
\label{fig:length_generalization}
\end{figure}

\paragraph{Generalization to Longer Context.}
Although our RL training caps the input length at $64$K tokens, the resulting gains transfer to substantially longer contexts. We re-evaluate Qwen3-4B-Thinking-2507 on LongBench~v2 with the maximum input length extended to $230$K tokens and stratify accuracy (avg@4) by example token length. As shown in the left panel of Figure~\ref{fig:length_generalization}, our model improves over the base by $+8.6$ on the in-distribution $0$--$64$K bucket and retains positive gains on the longer buckets, including $64$--$256$K ($+6.9$), $256$--$512$K ($+4.6$), and $512$K$+$ ($+2.1$), much beyond the training context. We further evaluate on HELMET~\citep{yen2025helmet}, which supports evaluation at multiple input lengths; scores at $32$K, $64$K, and $128$K are shown in the right panel of Figure~\ref{fig:length_generalization}. Together, these results indicate that our recipe instills a length-generic long-context reasoning skill rather than overfitting the $64$K training distribution.

\begin{table*}[!t]
\centering
\small
\setlength{\tabcolsep}{4pt}
\renewcommand{\arraystretch}{1.1}
\resizebox{\textwidth}{!}{%
\begin{tabular}{l c cc ccc cc}
\toprule
\multirow{2}{*}{\textbf{Data Category}} & \multirow{2}{*}{\textbf{Avg.}} & \multicolumn{2}{c}{\textbf{Multi-hop QA}} & \multicolumn{3}{c}{\textbf{Holistic LC Reasoning}} & \multicolumn{2}{c}{\textbf{Synthetic LC Reasoning}} \\
\cmidrule(lr){3-4} \cmidrule(lr){5-7} \cmidrule(lr){8-9}
 & & LBv1-QA & FRAMES & LBv2 & AA-LCR & DocFinQA & LongReason & GraphWalks \\
\midrule
Qwen3-4B-Thinking-2507                      & 53.42          & 62.60                            & 77.55                            & 40.60                            & 35.25                            & 39.80                            & 69.50                            & 48.67                            \\
\midrule
\quad + Retrieval                           & 55.83          & \cellcolor{gainbg}65.20          & \cellcolor{gainbg}\textbf{79.98} & \cellcolor{gainbg}44.78          & 32.25                            & \cellcolor{gainbg}48.70          & \cellcolor{gainbg}74.06          & 45.86                            \\
\quad + Multi-evidence                      & 56.44          & \cellcolor{gainbg}66.50          & \cellcolor{gainbg}77.79          & \cellcolor{gainbg}42.84          & 35.00                            & \cellcolor{gainbg}48.92          & \cellcolor{gainbg}73.68          & \cellcolor{gainbg}50.32          \\
\quad + Reasoning                           & 56.88          & \cellcolor{gainbg}65.00          & \cellcolor{gainbg}78.40          & \cellcolor{gainbg}44.73          & \cellcolor{gainbg}36.75          & \cellcolor{gainbg}48.05          & \cellcolor{gainbg}\textbf{75.06} & \cellcolor{gainbg}50.14          \\
\midrule
\quad + Retrieval\,+\,Multi-evidence        & 56.50          & \cellcolor{gainbg}67.00          & 77.18                            & \cellcolor{gainbg}41.45          & \cellcolor{gainbg}35.75          & \cellcolor{gainbg}\textbf{49.46} & \cellcolor{gainbg}74.31          & \cellcolor{gainbg}50.37          \\
\quad + Retrieval\,+\,Reasoning             & 57.48          & \cellcolor{gainbg}65.90          & \cellcolor{gainbg}78.16          & \cellcolor{gainbg}\textbf{45.58} & \cellcolor{gainbg}\textbf{37.25} & \cellcolor{gainbg}49.35          & \cellcolor{gainbg}74.94          & \cellcolor{gainbg}51.20          \\
\quad + Multi-evidence\,+\,Reasoning        & 55.98          & \cellcolor{gainbg}65.80          & 75.97                            & \cellcolor{gainbg}43.64          & \cellcolor{gainbg}36.50          & \cellcolor{gainbg}47.51          & \cellcolor{gainbg}72.42          & \cellcolor{gainbg}50.05          \\
\midrule
\quad + \textbf{All}                        & \textbf{57.58} & \cellcolor{gainbg}\textbf{67.70} & \cellcolor{gainbg}77.91          & \cellcolor{gainbg}44.63          & \cellcolor{gainbg}\textbf{37.25} & \cellcolor{gainbg}\textbf{49.46} & \cellcolor{gainbg}74.81          & \cellcolor{gainbg}\textbf{51.33} \\
\bottomrule
\end{tabular}%
}
\caption{Ablation on task categories of our data recipe. Each ``+\,$X$'' row trains the base model with only the data of the corresponding category, pairwise union, or the full union (``All''). \colorbox{gainbg}{Shaded cells} mark improvements over the base model on each individual benchmark (the Avg.\ column is left unshaded since all subsets improve the average). Within each column, \textbf{bold} marks the best score across the ablation rows.}
\label{tab:data_ablation}
\end{table*}

\section{Analysis}
We present ablation studies to assess the effect of task categories, task-balancing techniques, and reward design on long-context RL performance. Each ablation uses Qwen3-4B-Thinking-2507 with the same configuration as in Table~\ref{tab:main_results} except for the ablated component, and train for a fixed $50$ steps to ensure a fair comparison. We also apply our recipe to an agent-tuned model and show that the improved long-context reasoning benefits agentic performance as well.

\subsection{Ablation Study on Datasets}
To measure the contribution of each task category in our data recipe, we train Qwen3-4B-Thinking-2507 on each of the three task categories in isolation, as well as on every pairwise union, and compare them against the full mixture. The results are reported in Table~\ref{tab:data_ablation}.
Every single category and pairwise combination improves both the average and most individual benchmarks (marked by shaded cells in Table~\ref{tab:data_ablation}), confirming the effectiveness of our task categorization.
Among the three single-category subsets, \textbf{Reasoning} yields the largest gain ($+3.46$ on average), especially on LongBench~v2 ($+4.13$) and LongReason ($+5.56$), suggesting that training on long-form reasoning tasks is an effective method for long-context RL. The largest pairwise union improvement comes from \textbf{Retrieval\,+\,Reasoning}, which further improves LongBench~v2 and AA-LCR over Reasoning alone.
In addition, no single category or pairwise combination matches the full mixture (\textbf{All}, $57.58$), which indicates that the three task categories in our taxonomy are complementary, and all three are needed for the best result.

\subsection{Ablation on Task-balancing Techniques}
To assess the contribution of task-balanced sampling and task-level advantage normalization, we ablate both components and fall back to vanilla group-level normalization (\textbf{w/o task balancing} in Table~\ref{tab:rl_ablation}). Removing task balancing leads to a moderate drop in performance, with the average score decreasing from $57.58$ to $56.05$ (-$1.53$). This indicates that multi-task long-context RL still suffers from inter-task competition, and that task balancing is necessary to mitigate this effect. Nevertheless, the ablated variant still delivers consistent improvements over the base model on most benchmarks, suggesting the robustness of our data recipe to the choice of advantage normalization.

\begin{table*}[!t]
\centering
\small
\setlength{\tabcolsep}{4pt}
\renewcommand{\arraystretch}{1.1}
\resizebox{\textwidth}{!}{%
\begin{tabular}{l c cc ccc cc}
\toprule
\multirow{2}{*}{\textbf{Configuration}} & \multirow{2}{*}{\textbf{Avg.}} & \multicolumn{2}{c}{\textbf{Multi-hop QA}} & \multicolumn{3}{c}{\textbf{Holistic LC Reasoning}} & \multicolumn{2}{c}{\textbf{Synthetic LC Reasoning}} \\
\cmidrule(lr){3-4} \cmidrule(lr){5-7} \cmidrule(lr){8-9}
 & & LBv1-QA & FRAMES & LBv2 & AA-LCR & DocFinQA & LongReason & GraphWalks \\
\midrule
Qwen3-4B-Thinking-2507                & 53.42          & 62.60                            & 77.55                            & 40.60                            & 35.25                            & 39.80                            & 69.50                            & 48.67                            \\
\midrule
\quad + \textbf{Full recipe}          & \textbf{57.58} & \textbf{67.70} & \textbf{77.91} & 44.63          & \textbf{37.25} & \textbf{49.46} & \textbf{74.81} & \textbf{51.33} \\
\quad\quad w/o task balancing         & 56.05          & 64.80          & 76.58          & 42.15          & 35.00          & 48.59          & 74.06          & 51.16          \\
\quad\quad w/ process reward          & 56.29          & 65.90          & 75.97          & \textbf{44.83} & 35.00          & 48.16          & 73.55          & 50.64          \\
\bottomrule
\end{tabular}%
}
\caption{Ablation on RL design choices, all trained on top of Qwen3-4B-Thinking-2507 with our full data recipe. \textbf{Full recipe} reuses the configuration from Table~\ref{tab:main_results}. \textbf{w/o task balancing} disables both task-balanced sampling and task-level advantage normalization (\S\ref{sec:rl}). \textbf{w/ process reward} adds a thinking-process reward on top of the outcome reward. Within each column, \textbf{bold} marks the best score across the three configurations.}
\label{tab:rl_ablation}
\end{table*}

\subsection{Ablation on Reward Design}
A line of long-context RL work supplements the outcome reward with auxiliary signals over the model's thinking trajectory, often relying on additional evidence annotations~\citep{chen2026longrlvr,guan2026eapo}.
Since our training mixture contains no such annotations, we instead test a generic LLM-as-judge process reward that scores whether the thinking trajectory is internally coherent and correctly synthesizes the retrieved information into the final answer. We use DeepSeek-V3.2 as the judge and add this reward on top of our outcome-based reward.
As reported in Table~\ref{tab:rl_ablation} (\textbf{w/ process reward}), this auxiliary signal does not improve over the full recipe: the average score drops from $57.58$ to $56.29$ ($-1.29$).
We hypothesize two reasons. First, without ground-truth evidence, our LLM-as-judge can only score thinking coherence based on its own priors, yielding a noisier reward signal than evidence-anchored alternatives. Second, given that our diverse data recipe already provides an informative outcome signal, additional process-level supervision offers limited  benefit.

\subsection{Long-context Reasoning Improves Agentic Performance}

\begin{table}[t]
    \centering
    \small
    \setlength{\tabcolsep}{4pt}
    \renewcommand{\arraystretch}{1.15}
    \begin{tabular}{c c c c}
    \toprule
    \textbf{Model} & \textbf{GAIA} & \textbf{BrowseComp} & \textbf{Avg. LC} \\
    \midrule
    AgentCPM-Explore      & 66.1          & 27.00          & 45.57 \\
    \midrule
    \;+ Ours (25 steps)   & 67.7         & 29.00          & 52.92 \\
    \;+ Ours (50 steps)   & \textbf{70.9} & \textbf{34.00} & \textbf{54.64} \\
    \bottomrule
    \end{tabular}
    \caption{We train AgentCPM-Explore with our long-context data recipe and observe improvements on GAIA and BrowseComp (Pass@3) and on the long-context average score (\textbf{Avg.~LC}, over seven benchmarks). \textbf{Bold} marks the best score.}
    \label{tab:agent}
    \end{table}

Long-horizon agents naturally accumulate massive amounts of context across many turns of tool use, and recent analyses identify long-context capability as a bottleneck on their performance~\citep{li2026genagentbench}. Motivated by these observations, we conduct a preliminary investigation into whether the skills acquired through our long-context RL benefit long-horizon agentic tasks.

We conduct long-context RL training on AgentCPM-Explore~\citep{AgentCPMExplore2026}, an agent-enhanced model built on top of Qwen3-4B-Thinking-2507~\citep{qwen3}, with our proposed recipe and the same RL configuration as in the main experiments. 
In addition to the seven long-context benchmarks, we run evaluation on two long-horizon agent benchmarks, GAIA~\citep{mialon2023gaia} and BrowseComp~\citep{wei2025browsecomp}, both of which involve web search and frequently require the agent to process lengthy retrieved documents within a single session.
To prevent answer leakage~\citep{han2025stc}, we apply a rigorous filtering protocol; full evaluation details are described in Appendix~\ref{sec:agent_eval}. 

According to Table~\ref{tab:agent}, our long-context RL data translates into stronger agentic behavior beyond long-context reasoning gains: the $50$-step checkpoint improves GAIA Pass@3 by $+4.8$ and BrowseComp by $+7.0$ points over AgentCPM-Explore.
These preliminary results indicate that strengthening a model's ability to reason over long contexts can yield transferable gains to agentic scenarios.

\section{Conclusion}
We introduce a simple yet effective data recipe for long-context RL. By constructing a training mixture that contains three complementary tasks, which are retrieval, multi-evidence synthesis, and reasoning, we achieve consistent improvements across seven long-context reasoning benchmarks. Trained with our data recipe, Qwen3-4B-Thinking-2507, -8B, and -30B-A3B-Thinking-2507 reach average scores of $60.63$, $53.10$, and $68.23$, gaining $+7.21$, $+3.17$, and $+6.41$ over their respective base models. An intriguing finding is that these long-context gains transfer to agentic tasks: continuing RL training of an agent-tuned model on our data improves GAIA by $+4.8$ and BrowseComp by $+7.0$ points, indicating that strengthening core long-context abilities is an effective pathway toward stronger agentic capability.

\section*{Limitations}
Our experiments are limited to the Qwen3 family at scales up to 30B-A3B-Thinking, and we have not validated the recipe on larger models from other families due to computational constraints. We constrain all RL training to inputs under $64$K tokens, leaving extension to longer context lengths for future work. Most datasets in our mixture are synthetic; although they prove effective for improving long-context reasoning, they may not fully match the distribution of real-world long-context inputs. Our agentic transfer study is preliminary, constrained to one agent-tuned model and two benchmarks, and the broader relationship between long-context reasoning and agentic performance remains to be explored.

\section*{Ethics Statement}
This work does not involve individual information or offensive content. All training data is either synthesized or sourced from publicly available datasets, and our released data will follow the original licenses of the sources. 


\FloatBarrier

\bibliography{custom}

\begin{thebibliography}{58}
\providecommand{\natexlab}[1]{#1}

\bibitem[{Bai et~al.(2024{\natexlab{a}})Bai, Lv, Zhang, He, Qi, Hou, Tang, Dong, and Li}]{bai2024longalign}
Yushi Bai, Xin Lv, Jiajie Zhang, Yuze He, Ji~Qi, Lei Hou, Jie Tang, Yuxiao Dong, and Juanzi Li. 2024{\natexlab{a}}.
\newblock \href {https://doi.org/10.18653/v1/2024.findings-emnlp.74} {{L}ong{A}lign: A recipe for long context alignment of large language models}.
\newblock In \emph{Findings of the Association for Computational Linguistics: EMNLP 2024}, pages 1376--1395, Miami, Florida, USA. Association for Computational Linguistics.

\bibitem[{Bai et~al.(2024{\natexlab{b}})Bai, Lv, Zhang, Lyu, Tang, Huang, Du, Liu, Zeng, Hou, Dong, Tang, and Li}]{bai2024longbench}
Yushi Bai, Xin Lv, Jiajie Zhang, Hongchang Lyu, Jiankai Tang, Zhidian Huang, Zhengxiao Du, Xiao Liu, Aohan Zeng, Lei Hou, Yuxiao Dong, Jie Tang, and Juanzi Li. 2024{\natexlab{b}}.
\newblock \href {https://doi.org/10.18653/v1/2024.acl-long.172} {{L}ong{B}ench: A bilingual, multitask benchmark for long context understanding}.
\newblock In \emph{Proceedings of the 62nd Annual Meeting of the Association for Computational Linguistics (Volume 1: Long Papers)}, pages 3119--3137, Bangkok, Thailand. Association for Computational Linguistics.

\bibitem[{Bai et~al.(2025)Bai, Tu, Zhang, Peng, Wang, Lv, Cao, Xu, Hou, Dong, Tang, and Li}]{bai2025longbenchv2}
Yushi Bai, Shangqing Tu, Jiajie Zhang, Hao Peng, Xiaozhi Wang, Xin Lv, Shulin Cao, Jiazheng Xu, Lei Hou, Yuxiao Dong, Jie Tang, and Juanzi Li. 2025.
\newblock \href {https://doi.org/10.18653/v1/2025.acl-long.183} {{L}ong{B}ench v2: Towards deeper understanding and reasoning on realistic long-context multitasks}.
\newblock In \emph{Proceedings of the 63rd Annual Meeting of the Association for Computational Linguistics (Volume 1: Long Papers)}, pages 3639--3664, Vienna, Austria. Association for Computational Linguistics.

\bibitem[{Chen et~al.(2026{\natexlab{a}})Chen, Shieh, and Bing}]{chen2026longrlvr}
Guanzheng Chen, Michael~Qizhe Shieh, and Lidong Bing. 2026{\natexlab{a}}.
\newblock \href {https://arxiv.org/abs/2603.02146} {Longrlvr: Long-context reinforcement learning requires verifiable context rewards}.
\newblock \emph{Preprint}, arXiv:2603.02146.

\bibitem[{Chen et~al.(2026{\natexlab{b}})Chen, Cong, Fan, Fu, Gong, Lu, Li, Niu, Pan, Song, Wang, Wu, Wu, Xie, Yan, Zhang, Lin, Liu, and Sun}]{AgentCPMExplore2026}
Haotian Chen, Xin Cong, Shengda Fan, Yuyang Fu, Ziqin Gong, Yaxi Lu, Yishan Li, Boye Niu, Chengjun Pan, Zijun Song, Huadong Wang, Yesai Wu, Yueying Wu, Zihao Xie, Yukun Yan, Zhong Zhang, Yankai Lin, Zhiyuan Liu, and Maosong Sun. 2026{\natexlab{b}}.
\newblock \href {https://arxiv.org/abs/2602.06485} {Agentcpm-explore: Realizing long-horizon deep exploration for edge-scale agents}.
\newblock \emph{Preprint}, arXiv:2602.06485.

\bibitem[{Dasigi et~al.(2021)Dasigi, Lo, Beltagy, Cohan, Smith, and Gardner}]{dasigi2021qasper}
Pradeep Dasigi, Kyle Lo, Iz~Beltagy, Arman Cohan, Noah~A. Smith, and Matt Gardner. 2021.
\newblock \href {https://doi.org/10.18653/v1/2021.naacl-main.365} {A dataset of information-seeking questions and answers anchored in research papers}.
\newblock In \emph{Proceedings of the 2021 Conference of the North American Chapter of the Association for Computational Linguistics: Human Language Technologies}, pages 4599--4610, Online. Association for Computational Linguistics.

\bibitem[{Du et~al.(2025)Du, Tian, Ronanki, Rongali, Bodapati, Galstyan, Wells, Schwartz, Huerta, and Peng}]{du2025contextlength}
Yufeng Du, Minyang Tian, Srikanth Ronanki, Subendhu Rongali, Sravan~Babu Bodapati, Aram Galstyan, Azton Wells, Roy Schwartz, Eliu~A Huerta, and Hao Peng. 2025.
\newblock \href {https://doi.org/10.18653/v1/2025.findings-emnlp.1264} {Context length alone hurts {LLM} performance despite perfect retrieval}.
\newblock In \emph{Findings of the Association for Computational Linguistics: EMNLP 2025}, pages 23281--23298, Suzhou, China. Association for Computational Linguistics.

\bibitem[{Guan et~al.(2026)Guan, Li, Huang, Xie, Zhou, and Cao}]{guan2026eapo}
Xin Guan, Zijian Li, Shen Huang, Pengjun Xie, Jingren Zhou, and Jiuxin Cao. 2026.
\newblock \href {https://arxiv.org/abs/2601.10306} {Evidence-augmented policy optimization with reward co-evolution for long-context reasoning}.
\newblock \emph{Preprint}, arXiv:2601.10306.

\bibitem[{Guo et~al.(2025)Guo, Yang, Zhang, Song, Wang, Zhu, Xu, Zhang, Ma, Bi, Zhang, Yu, Wu, Wu, Gou, Shao, Li, Gao, Liu, Xue, Wang, Wu, Feng, Lu, Zhao, Deng, Ruan, Dai, Chen, Ji, Li, Lin, Dai, Luo, Hao, Chen, Li, Zhang, Xu, Ding, Gao, Qu, Li, Guo, Li, Chen, Yuan, Tu, Qiu, Li, Cai, Ni, Liang, Chen, Dong, Hu, You, Gao, Guan, Huang, Yu, Wang, Zhang, Zhao, Wang, Zhang, Xu, Xia, Zhang, Zhang, Tang, Zhou, Li, Wang, Li, Tian, Huang, Zhang, Wang, Chen, Du, Ge, Zhang, Pan, Wang, Chen, Jin, Chen, Lu, Zhou, Chen, Ye, Wang, Yu, Zhou, Pan, Li, Zhou, Wu, Yun, Pei, Sun, Wang, Zeng, Liu, Liang, Gao, Yu, Zhang, Xiao, An, Liu, Wang, Chen, Nie, Cheng, Liu, Xie, Liu, Yang, Li, Su, Lin, Li, Jin, Shen, Chen, Sun, Wang, Song, Zhou, Wang, Shan, Li, Wang, Wei, Zhang, Xu, Li, Zhao, Sun, Wang, Yu, Zhang, Shi, Xiong, He, Piao, Wang, Tan, Ma, Liu, Guo, Ou, Wang, Gong, Zou, He, Xiong, Luo, You, Liu, Zhou, Zhu, Huang, Li, Zheng, Zhu, Ma, Tang, Zha, Yan, Ren, Ren, Sha, Fu, Xu, Xie, Zhang, Hao, Ma, Yan, Wu, Gu, Zhu, Liu, Li, Xie, Song,
  Pan, Huang, Xu, Zhang, and Zhang}]{deepseekai2025r1}
Daya Guo, Dejian Yang, Haowei Zhang, Junxiao Song, Peiyi Wang, Qihao Zhu, Runxin Xu, Ruoyu Zhang, Shirong Ma, Xiao Bi, Xiaokang Zhang, Xingkai Yu, Yu~Wu, Z.~F. Wu, Zhibin Gou, Zhihong Shao, Zhuoshu Li, Ziyi Gao, Aixin Liu, and 175 others. 2025.
\newblock \href {https://doi.org/10.1038/s41586-025-09422-z} {Deepseek-r1 incentivizes reasoning in llms through reinforcement learning}.
\newblock \emph{Nature}, 645(8081):633–638.

\bibitem[{Gupta et~al.(2025)Gupta, Zhu, Sharma, O{'}Brien, and Lu}]{gupta2025novelhopqa}
Abhay Gupta, Kevin Zhu, Vasu Sharma, Sean O{'}Brien, and Michael Lu. 2025.
\newblock \href {https://doi.org/10.18653/v1/2025.emnlp-main.1328} {{N}ovel{H}op{QA}: Diagnosing multi-hop reasoning failures in long narrative contexts}.
\newblock In \emph{Proceedings of the 2025 Conference on Empirical Methods in Natural Language Processing}, pages 26134--26151, Suzhou, China. Association for Computational Linguistics.

\bibitem[{Han et~al.(2025)Han, Mankikar, Michael, and Wang}]{han2025stc}
Ziwen Han, Meher Mankikar, Julian Michael, and Zifan Wang. 2025.
\newblock \href {https://openreview.net/forum?id=FBoIUyY3wX} {Search-time data contamination}.
\newblock In \emph{NeurIPS 2025 Workshop on Evaluating the Evolving LLM Lifecycle: Benchmarks, Emergent Abilities, and Scaling}.

\bibitem[{He et~al.(2026)He, Liang, Xu, Liu, Chen, Wang, Song, Yu, Liang, Wang, Zhang, Wang, Tu, Mi, and Yu}]{he2025deepmath}
Zhiwei He, Tian Liang, Jiahao Xu, Qiuzhi Liu, Xingyu Chen, Yue Wang, Linfeng Song, Dian Yu, Zhenwen Liang, Wenxuan Wang, Zhuosheng Zhang, Rui Wang, Zhaopeng Tu, Haitao Mi, and Dong Yu. 2026.
\newblock \href {https://openreview.net/forum?id=kHB5Te5IWm} {Deepmath-103k: A large-scale, challenging, decontaminated, and verifiable mathematical dataset for advancing reasoning}.
\newblock In \emph{The Fourteenth International Conference on Learning Representations}.

\bibitem[{Hendrycks et~al.(2021)Hendrycks, Burns, Kadavath, Arora, Basart, Tang, Song, and Steinhardt}]{hendrycks2021math}
Dan Hendrycks, Collin Burns, Saurav Kadavath, Akul Arora, Steven Basart, Eric Tang, Dawn Song, and Jacob Steinhardt. 2021.
\newblock \href {https://openreview.net/forum?id=7Bywt2mQsCe} {Measuring mathematical problem solving with the {MATH} dataset}.
\newblock In \emph{Thirty-fifth Conference on Neural Information Processing Systems Datasets and Benchmarks Track (Round 2)}.

\bibitem[{Hsieh et~al.(2024)Hsieh, Sun, Kriman, Acharya, Rekesh, Jia, and Ginsburg}]{hsieh2024ruler}
Cheng-Ping Hsieh, Simeng Sun, Samuel Kriman, Shantanu Acharya, Dima Rekesh, Fei Jia, and Boris Ginsburg. 2024.
\newblock \href {https://openreview.net/forum?id=kIoBbc76Sy} {{RULER}: What{\textquoteright}s the real context size of your long-context language models?}
\newblock In \emph{First Conference on Language Modeling}.

\bibitem[{Kamradt(2023)}]{kamradt2023niah}
Greg Kamradt. 2023.
\newblock Needle in a haystack -- pressure testing {LLM}s.
\newblock \url{https://github.com/gkamradt/LLMTest_NeedleInAHaystack}.

\bibitem[{Krishna et~al.(2025)Krishna, Krishna, Mohananey, Schwarcz, Stambler, Upadhyay, and Faruqui}]{gemini2024frames}
Satyapriya Krishna, Kalpesh Krishna, Anhad Mohananey, Steven Schwarcz, Adam Stambler, Shyam Upadhyay, and Manaal Faruqui. 2025.
\newblock \href {https://doi.org/10.18653/v1/2025.naacl-long.243} {Fact, fetch, and reason: A unified evaluation of retrieval-augmented generation}.
\newblock In \emph{Proceedings of the 2025 Conference of the Nations of the Americas Chapter of the Association for Computational Linguistics: Human Language Technologies (Volume 1: Long Papers)}, pages 4745--4759, Albuquerque, New Mexico. Association for Computational Linguistics.

\bibitem[{Levy et~al.(2024)Levy, Jacoby, and Goldberg}]{levy2024sametask}
Mosh Levy, Alon Jacoby, and Yoav Goldberg. 2024.
\newblock \href {https://doi.org/10.18653/v1/2024.acl-long.818} {Same task, more tokens: the impact of input length on the reasoning performance of large language models}.
\newblock In \emph{Proceedings of the 62nd Annual Meeting of the Association for Computational Linguistics (Volume 1: Long Papers)}, pages 15339--15353, Bangkok, Thailand. Association for Computational Linguistics.

\bibitem[{Li et~al.(2024{\natexlab{a}})Li, Verga, Sen, Yang, Viswanathan, Lewis, Watanabe, and Su}]{li2024alr2}
Huayang Li, Pat Verga, Priyanka Sen, Bowen Yang, Vijay Viswanathan, Patrick Lewis, Taro Watanabe, and Yixuan Su. 2024{\natexlab{a}}.
\newblock \href {https://arxiv.org/abs/2410.03227} {Alr$^2$: A retrieve-then-reason framework for long-context question answering}.
\newblock \emph{Preprint}, arXiv:2410.03227.

\bibitem[{Li et~al.(2025)Li, Zhang, Zhang, Duan, Liu, and Chen}]{li2024needlebench}
Mo~Li, Songyang Zhang, Taolin Zhang, Haodong Duan, Yunxin Liu, and Kai Chen. 2025.
\newblock \href {https://openreview.net/forum?id=cEvmIKsRw0} {Needlebench: Evaluating {LLM} retrieval and reasoning across varying information densities}.
\newblock \emph{Transactions on Machine Learning Research}.

\bibitem[{Li et~al.(2024{\natexlab{b}})Li, Yang, Cheng, Liu, Yu, Yang, and Lam}]{li2024selfimprovelong}
Siheng Li, Cheng Yang, Zesen Cheng, Lemao Liu, Mo~Yu, Yujiu Yang, and Wai Lam. 2024{\natexlab{b}}.
\newblock \href {https://arxiv.org/abs/2411.08147} {Large language models can self-improve in long-context reasoning}.
\newblock \emph{Preprint}, arXiv:2411.08147.

\bibitem[{Li et~al.(2026)Li, Ming, Setlur, Paladugu, Tang, Kang, Shao, Jin, and Xiong}]{li2026genagentbench}
Xiaochuan Li, Ryan Ming, Pranav Setlur, Abhijay Paladugu, Andy Tang, Hao Kang, Shuai Shao, Rong Jin, and Chenyan Xiong. 2026.
\newblock \href {https://arxiv.org/abs/2602.18998} {Benchmark test-time scaling of general llm agents}.
\newblock \emph{Preprint}, arXiv:2602.18998.

\bibitem[{Ling et~al.(2025)Ling, Liu, Yan, Yang, Lin, Fan, Shen, Du, and Chen}]{longreason}
Zhan Ling, Kang Liu, Kai Yan, Yifan Yang, Weijian Lin, Ting-Han Fan, Lingfeng Shen, Zhengyin Du, and Jiecao Chen. 2025.
\newblock \href {https://arxiv.org/abs/2501.15089} {Longreason: A synthetic long-context reasoning benchmark via context expansion}.
\newblock \emph{Preprint}, arXiv:2501.15089.

\bibitem[{Liu et~al.(2025)Liu, Li, Zhang, Li, Chen, Ji, Cheng, Wu, Du, Xu, Song, Zhu, Chen, Zhao, and He}]{liu2025webexplorer}
Junteng Liu, Yunji Li, Chi Zhang, Jingyang Li, Aili Chen, Ke~Ji, Weiyu Cheng, Zijia Wu, Chengyu Du, Qidi Xu, Jiayuan Song, Zhengmao Zhu, Wenhu Chen, Pengyu Zhao, and Junxian He. 2025.
\newblock \href {https://arxiv.org/abs/2509.06501} {Webexplorer: Explore and evolve for training long-horizon web agents}.
\newblock \emph{Preprint}, arXiv:2509.06501.

\bibitem[{Lu et~al.(2025)Lu, Hou, Wang, Zhang, Liu, Li, Feng, Tang, and Dong}]{lu2025deepdive}
Rui Lu, Zhenyu Hou, Zihan Wang, Hanchen Zhang, Xiao Liu, Yujiang Li, Shi Feng, Jie Tang, and Yuxiao Dong. 2025.
\newblock \href {https://arxiv.org/abs/2509.10446} {Deepdive: Advancing deep search agents with knowledge graphs and multi-turn rl}.
\newblock \emph{Preprint}, arXiv:2509.10446.

\bibitem[{Mialon et~al.(2024)Mialon, Fourrier, Wolf, LeCun, and Scialom}]{mialon2023gaia}
Gr{\'e}goire Mialon, Cl{\'e}mentine Fourrier, Thomas Wolf, Yann LeCun, and Thomas Scialom. 2024.
\newblock \href {https://openreview.net/forum?id=fibxvahvs3} {{GAIA}: a benchmark for general {AI} assistants}.
\newblock In \emph{The Twelfth International Conference on Learning Representations}.

\bibitem[{Modarressi et~al.(2025)Modarressi, Deilamsalehy, Dernoncourt, Bui, Rossi, Yoon, and Schuetze}]{modarressi2025nolima}
Ali Modarressi, Hanieh Deilamsalehy, Franck Dernoncourt, Trung Bui, Ryan~A. Rossi, Seunghyun Yoon, and Hinrich Schuetze. 2025.
\newblock \href {https://openreview.net/forum?id=0OshX1hiSa} {Nolima: Long-context evaluation beyond literal matching}.
\newblock In \emph{Forty-second International Conference on Machine Learning}.

\bibitem[{OpenAI et~al.(2026)OpenAI, :, Jaech, Kalai, Lerer, Richardson, El-Kishky, Low, Helyar, Madry, Beutel, Carney, Iftimie, Karpenko, Passos, Neitz, Prokofiev, Wei, Tam, Bennett, Kumar, Saraiva, Vallone, Duberstein, Kondrich, Mishchenko, Applebaum, Jiang, Nair, Zoph, Ghorbani, Zhang, Rossen, Sokolowsky, Barak, McGrew, Minaiev, Hao, Baker, Houghton, McKinzie, Eastman, Lugaresi, Bassin, Hudson, Li, de~Bourcy, Voss, Shen, Zhang, Koch, Orsinger, Hesse, Fischer, Chan, Roberts, Kappler, Levy, Selsam, Dohan, Farhi, Mely, Robinson, Tsipras, Li, Oprica, Freeman, Zhang, Wong, Proehl, Cheung, Mitchell, Wallace, Ritter, Mays, Wang, Such, Raso, Leoni, Tsimpourlas, Song, von Lohmann, Sulit, Salmon, Parascandolo, Chabot, Zhao, Brockman, Leclerc, Salman, Bao, Sheng, Andrin, Bagherinezhad, Ren, Lightman, Chung, Kivlichan, O'Connell, Osband, Gilaberte, Akkaya, Kostrikov, Sutskever, Kofman, Pachocki, Lennon, Wei, Harb, Twore, Feng, Yu, Weng, Tang, Yu, Candela, Palermo, Parish, Heidecke, Hallman, Rizzo, Gordon, Uesato,
  Ward, Huizinga, Wang, Chen, Xiao, Singhal, Nguyen, Cobbe, Shi, Wood, Rimbach, Gu-Lemberg, Liu, Lu, Stone, Yu, Ahmad, Yang, Liu, Maksin, Ho, Fedus, Weng, Li, McCallum, Held, Kuhn, Kondraciuk, Kaiser, Metz, Boyd, Trebacz, Joglekar, Chen, Tintor, Meyer, Jones, Kaufer, Schwarzer, Shah, Yatbaz, Guan, Xu, Yan, Glaese, Chen, Lampe, Malek, Wang, Fradin, McClay, Pavlov, Wang, Wang, Murati, Bavarian, Rohaninejad, McAleese, Chowdhury, Chowdhury, Ryder, Tezak, Brown, Nachum, Boiko, Murk, Watkins, Chao, Ashbourne, Izmailov, Zhokhov, Dias, Arora, Lin, Lopes, Gaon, Miyara, Leike, Hwang, Garg, Brown, James, Shu, Cheu, Greene, Jain, Altman, Toizer, Toyer, Miserendino, Agarwal, Hernandez, Baker, McKinney, Yan, Zhao, Hu, Santurkar, Chaudhuri, Zhang, Fu, Papay, Lin, Balaji, Sanjeev, Sidor, Broda, Clark, Wang, Gordon, Sanders, Patwardhan, Sottiaux, Degry, Dimson, Zheng, Garipov, Stasi, Bansal, Creech, Peterson, Eloundou, Qi, Kosaraju, Monaco, Pong, Fomenko, Zheng, Zhou, Zhan, McCabe, Zaremba, Dubois, Lu, Chen, Cha, Bai, He,
  Zhang, Wang, Shao, and Li}]{openai2024o1}
OpenAI, :, Aaron Jaech, Adam Kalai, Adam Lerer, Adam Richardson, Ahmed El-Kishky, Aiden Low, Alec Helyar, Aleksander Madry, Alex Beutel, Alex Carney, Alex Iftimie, Alex Karpenko, Alex~Tachard Passos, Alexander Neitz, Alexander Prokofiev, Alexander Wei, Allison Tam, and 246 others. 2026.
\newblock \href {https://arxiv.org/abs/2412.16720} {Openai o1 system card}.
\newblock \emph{Preprint}, arXiv:2412.16720.

\bibitem[{{OpenAI}(2025)}]{graphwalks}
{OpenAI}. 2025.
\newblock {GraphWalks}: a multi hop reasoning long context benchmark.
\newblock \url{https://huggingface.co/datasets/openai/graphwalks}.
\newblock Released alongside GPT-4.1 (April 2025); see \url{https://openai.com/index/gpt-4-1/}.

\bibitem[{Peng et~al.(2024)Peng, Quesnelle, Fan, and Shippole}]{peng2023yarn}
Bowen Peng, Jeffrey Quesnelle, Honglu Fan, and Enrico Shippole. 2024.
\newblock \href {https://openreview.net/forum?id=wHBfxhZu1u} {Ya{RN}: Efficient context window extension of large language models}.
\newblock In \emph{The Twelfth International Conference on Learning Representations}.

\bibitem[{Peng et~al.(2026)Peng, Shen, Chen, Li, Yan, and Li}]{peng2026longpas}
Miao Peng, Weizhou Shen, Nuo Chen, Chenliang Li, Ming Yan, and Jia Li. 2026.
\newblock \href {https://arxiv.org/abs/2601.12465} {Incentivizing in-depth reasoning over long contexts with process advantage shaping}.
\newblock \emph{Preprint}, arXiv:2601.12465.

\bibitem[{Ping et~al.(2026{\natexlab{a}})Ping, Chen, Hui, Yu, Li, Yan, and Chang}]{ping2026longact}
Bowen Ping, Zijun Chen, Tingfeng Hui, Qize Yu, Chenxuan Li, Junchi Yan, and Baobao Chang. 2026{\natexlab{a}}.
\newblock \href {https://arxiv.org/abs/2604.14922} {Longact: Harnessing intrinsic activation patterns for long-context reinforcement learning}.
\newblock \emph{Preprint}, arXiv:2604.14922.

\bibitem[{Ping et~al.(2026{\natexlab{b}})Ping, Chen, Yu, Hui, Yan, and Chang}]{ping2026longr}
Bowen Ping, Zijun Chen, Yiyao Yu, Tingfeng Hui, Junchi Yan, and Baobao Chang. 2026{\natexlab{b}}.
\newblock \href {https://arxiv.org/abs/2602.05758} {Longr: Unleashing long-context reasoning via reinforcement learning with dense utility rewards}.
\newblock \emph{Preprint}, arXiv:2602.05758.

\bibitem[{Qin et~al.(2025)Qin, Ye, Fang, Wang, Liang, Tian, Zhang, Li, Li, Huang, Zhong, Li, Yang, Miao, Lin, Liu, Jiang, Ma, Li, Xiao, Cai, Li, Zheng, Jin, Li, Zhou, Wang, Chen, Li, Yang, Liu, Lin, Peng, Liu, and Shi}]{qin2025uitars}
Yujia Qin, Yining Ye, Junjie Fang, Haoming Wang, Shihao Liang, Shizuo Tian, Junda Zhang, Jiahao Li, Yunxin Li, Shijue Huang, Wanjun Zhong, Kuanye Li, Jiale Yang, Yu~Miao, Woyu Lin, Longxiang Liu, Xu~Jiang, Qianli Ma, Jingyu Li, and 16 others. 2025.
\newblock \href {https://arxiv.org/abs/2501.12326} {Ui-tars: Pioneering automated gui interaction with native agents}.
\newblock \emph{Preprint}, arXiv:2501.12326.

\bibitem[{{Qwen Team}(2025{\natexlab{a}})}]{qwen3-30b-a3b-thinking-2507}
{Qwen Team}. 2025{\natexlab{a}}.
\newblock {Qwen3-30B-A3B-Thinking-2507} model card.
\newblock \url{https://huggingface.co/Qwen/Qwen3-30B-A3B-Thinking-2507}.
\newblock Accessed: 2026-05-25.

\bibitem[{{Qwen Team}(2025{\natexlab{b}})}]{qwen3-4b-thinking-2507}
{Qwen Team}. 2025{\natexlab{b}}.
\newblock {Qwen3-4B-Thinking-2507} model card.
\newblock \url{https://huggingface.co/Qwen/Qwen3-4B-Thinking-2507}.
\newblock Accessed: 2026-05-25.

\bibitem[{{Qwen Team}(2025{\natexlab{c}})}]{qwen3-8b-card}
{Qwen Team}. 2025{\natexlab{c}}.
\newblock {Qwen3-8B} model card.
\newblock \url{https://huggingface.co/Qwen/Qwen3-8B}.
\newblock Accessed: 2026-05-25.

\bibitem[{{Qwen Team}(2025{\natexlab{d}})}]{qwen3}
{Qwen Team}. 2025{\natexlab{d}}.
\newblock \href {https://arxiv.org/abs/2505.09388} {{Qwen3} technical report}.
\newblock \emph{arXiv preprint arXiv:2505.09388}.

\bibitem[{{RadixArk}(2025)}]{miles2025}
{RadixArk}. 2025.
\newblock Miles.
\newblock \url{https://github.com/radixark/miles}.
\newblock Miles is an enterprise-facing reinforcement learning framework for LLM and VLM post-training, forked from and co-evolving with slime.

\bibitem[{Reddy et~al.(2024)Reddy, Koncel-Kedziorski, Lai, Krumdick, Lovering, and Tanner}]{docfinqa}
Varshini Reddy, Rik Koncel-Kedziorski, Viet~Dac Lai, Michael Krumdick, Charles Lovering, and Chris Tanner. 2024.
\newblock \href {https://doi.org/10.18653/v1/2024.acl-short.42} {{D}oc{F}in{QA}: A long-context financial reasoning dataset}.
\newblock In \emph{Proceedings of the 62nd Annual Meeting of the Association for Computational Linguistics (Volume 2: Short Papers)}, pages 445--458, Bangkok, Thailand. Association for Computational Linguistics.

\bibitem[{Shao et~al.(2024)Shao, Wang, Zhu, Xu, Song, Bi, Zhang, Zhang, Li, Wu, and Guo}]{shao2024deepseekmath}
Zhihong Shao, Peiyi Wang, Qihao Zhu, Runxin Xu, Junxiao Song, Xiao Bi, Haowei Zhang, Mingchuan Zhang, Y.~K. Li, Y.~Wu, and Daya Guo. 2024.
\newblock \href {https://arxiv.org/abs/2402.03300} {Deepseekmath: Pushing the limits of mathematical reasoning in open language models}.
\newblock \emph{Preprint}, arXiv:2402.03300.

\bibitem[{Shen et~al.(2025)Shen, Yang, Li, Lu, Peng, Sun, Shi, Liao, Lai, Zhang, Liu, Huang, Zhou, and Yan}]{shen2025qwenlongl15}
Weizhou Shen, Ziyi Yang, Chenliang Li, Zhiyuan Lu, Miao Peng, Huashan Sun, Yingcheng Shi, Shengyi Liao, Shaopeng Lai, Bo~Zhang, Dayiheng Liu, Fei Huang, Jingren Zhou, and Ming Yan. 2025.
\newblock \href {https://arxiv.org/abs/2512.12967} {Qwenlong-l1.5: Post-training recipe for long-context reasoning and memory management}.
\newblock \emph{Preprint}, arXiv:2512.12967.

\bibitem[{Shoeybi et~al.(2020)Shoeybi, Patwary, Puri, LeGresley, Casper, and Catanzaro}]{shoeybi2019megatron}
Mohammad Shoeybi, Mostofa Patwary, Raul Puri, Patrick LeGresley, Jared Casper, and Bryan Catanzaro. 2020.
\newblock \href {https://arxiv.org/abs/1909.08053} {Megatron-lm: Training multi-billion parameter language models using model parallelism}.
\newblock \emph{Preprint}, arXiv:1909.08053.

\bibitem[{Su et~al.(2026)Su, Fang, Huang, Zeng, Zhao, Shi, Zhang, Chen, Chen, Wu, and Zhao}]{su2026acc}
Qisheng Su, Zhen Fang, Shiting Huang, Yu~Zeng, Yiming Zhao, Kou Shi, Ziao Zhang, Lin Chen, Zehui Chen, Lijun Wu, and Feng Zhao. 2026.
\newblock \href {https://arxiv.org/abs/2605.21850} {Acc: Compiling agent trajectories for long-context training}.
\newblock \emph{Preprint}, arXiv:2605.21850.

\bibitem[{Tang and Yang(2024)}]{tang2024multihoprag}
Yixuan Tang and Yi~Yang. 2024.
\newblock \href {https://openreview.net/forum?id=t4eB3zYWBK} {Multihop-{RAG}: Benchmarking retrieval-augmented generation for multi-hop queries}.
\newblock In \emph{First Conference on Language Modeling}.

\bibitem[{Team(2025)}]{aalcr2024}
Artificial~Analysis Team. 2025.
\newblock Artificial analysis long context reasoning benchmark(lcr).

\bibitem[{Trivedi et~al.(2022)Trivedi, Balasubramanian, Khot, and Sabharwal}]{trivedi2022musique}
Harsh Trivedi, Niranjan Balasubramanian, Tushar Khot, and Ashish Sabharwal. 2022.
\newblock \href {https://doi.org/10.1162/tacl_a_00475} {{M}u{S}i{Q}ue: Multihop questions via single-hop question composition}.
\newblock \emph{Transactions of the Association for Computational Linguistics}, 10:539--554.

\bibitem[{Vodrahalli et~al.(2024)Vodrahalli, Ontanon, Tripuraneni, Xu, Jain, Shivanna, Hui, Dikkala, Kazemi, Fatemi, Anil, Dyer, Shakeri, Vij, Mehta, Ramasesh, Le, Chi, Lu, Firat, Lazaridou, Lespiau, Attaluri, and Olszewska}]{vodrahalli2024mrcr}
Kiran Vodrahalli, Santiago Ontanon, Nilesh Tripuraneni, Kelvin Xu, Sanil Jain, Rakesh Shivanna, Jeffrey Hui, Nishanth Dikkala, Mehran Kazemi, Bahare Fatemi, Rohan Anil, Ethan Dyer, Siamak Shakeri, Roopali Vij, Harsh Mehta, Vinay Ramasesh, Quoc Le, Ed~Chi, Yifeng Lu, and 5 others. 2024.
\newblock \href {https://arxiv.org/abs/2409.12640} {Michelangelo: Long context evaluations beyond haystacks via latent structure queries}.
\newblock \emph{Preprint}, arXiv:2409.12640.

\bibitem[{Wan et~al.(2025)Wan, Shen, Liao, Shi, Li, Yang, Zhang, Huang, Zhou, and Yan}]{wan2025qwenlongl1}
Fanqi Wan, Weizhou Shen, Shengyi Liao, Yingcheng Shi, Chenliang Li, Ziyi Yang, Ji~Zhang, Fei Huang, Jingren Zhou, and Ming Yan. 2025.
\newblock \href {https://arxiv.org/abs/2505.17667} {Qwenlong-l1: Towards long-context large reasoning models with reinforcement learning}.
\newblock \emph{Preprint}, arXiv:2505.17667.

\bibitem[{Wang et~al.(2026)Wang, Zhang, Zhang, Shang, Yang, Chen, and Yang}]{wang2026loongrl}
Siyuan Wang, Gaokai Zhang, Li~Lyna Zhang, Ning Shang, Fan Yang, Dongyao Chen, and Mao Yang. 2026.
\newblock \href {https://openreview.net/forum?id=o29E01Q6bv} {Loong{RL}: Reinforcement learning for advanced reasoning over long contexts}.
\newblock In \emph{The Fourteenth International Conference on Learning Representations}.

\bibitem[{Wei et~al.(2025)Wei, Sun, Papay, McKinney, Han, Fulford, Chung, Passos, Fedus, and Glaese}]{wei2025browsecomp}
Jason Wei, Zhiqing Sun, Spencer Papay, Scott McKinney, Jeffrey Han, Isa Fulford, Hyung~Won Chung, Alex~Tachard Passos, William Fedus, and Amelia Glaese. 2025.
\newblock \href {https://arxiv.org/abs/2504.12516} {Browsecomp: A simple yet challenging benchmark for browsing agents}.
\newblock \emph{Preprint}, arXiv:2504.12516.

\bibitem[{Whitecross and Rahimi(2026)}]{whitecross2026recallm}
Kyle Whitecross and Negin Rahimi. 2026.
\newblock \href {https://arxiv.org/abs/2604.09494} {Recallm: Addressing the lost-in-thought phenomenon with explicit in-context retrieval}.
\newblock \emph{Preprint}, arXiv:2604.09494.

\bibitem[{Xiao et~al.(2026)Xiao, Xie, Zhao, Dou, Wang, Liu, Zheng, Zhang, Zhou, Zhang, and Liu}]{xiao2026decomposition}
Yanling Xiao, Huaibing Xie, Guoliang Zhao, Shihan Dou, Shaolei Wang, Yiting Liu, Nantao Zheng, Cheng Zhang, Pluto Zhou, Zhisong Zhang, and Lemao Liu. 2026.
\newblock \href {https://arxiv.org/abs/2604.07981} {A decomposition perspective to long-context reasoning for llms}.
\newblock \emph{Preprint}, arXiv:2604.07981.

\bibitem[{Xu et~al.(2024)Xu, Ye, and Ren}]{xu2024lifelongicl}
Xiaoyue Xu, Qinyuan Ye, and Xiang Ren. 2024.
\newblock \href {https://openreview.net/forum?id=5ltgEKXnZt} {Stress-testing long-context language models with lifelong {ICL} and task haystack}.
\newblock In \emph{First Workshop on Long-Context Foundation Models @ ICML 2024}.

\bibitem[{Yang et~al.(2024)Yang, Jimenez, Wettig, Lieret, Yao, Narasimhan, and Press}]{yang2024sweagent}
John Yang, Carlos~E Jimenez, Alexander Wettig, Kilian Lieret, Shunyu Yao, Karthik~R Narasimhan, and Ofir Press. 2024.
\newblock \href {https://openreview.net/forum?id=mXpq6ut8J3} {{SWE}-agent: Agent-computer interfaces enable automated software engineering}.
\newblock In \emph{The Thirty-eighth Annual Conference on Neural Information Processing Systems}.

\bibitem[{Yang et~al.(2018)Yang, Qi, Zhang, Bengio, Cohen, Salakhutdinov, and Manning}]{yang2018hotpotqa}
Zhilin Yang, Peng Qi, Saizheng Zhang, Yoshua Bengio, William Cohen, Ruslan Salakhutdinov, and Christopher~D. Manning. 2018.
\newblock \href {https://doi.org/10.18653/v1/D18-1259} {{H}otpot{QA}: A dataset for diverse, explainable multi-hop question answering}.
\newblock In \emph{Proceedings of the 2018 Conference on Empirical Methods in Natural Language Processing}, pages 2369--2380, Brussels, Belgium. Association for Computational Linguistics.

\bibitem[{Yen et~al.(2025)Yen, Gao, Hou, Ding, Fleischer, Izsak, Wasserblat, and Chen}]{yen2025helmet}
Howard Yen, Tianyu Gao, Minmin Hou, Ke~Ding, Daniel Fleischer, Peter Izsak, Moshe Wasserblat, and Danqi Chen. 2025.
\newblock \href {https://openreview.net/forum?id=293V3bJbmE} {{HELMET}: How to evaluate long-context models effectively and thoroughly}.
\newblock In \emph{The Thirteenth International Conference on Learning Representations}.

\bibitem[{Yu et~al.(2026)Yu, Zhang, Zhu, Yuan, Zuo, YuYue, Dai, Fan, Liu, Liu, Liu, Liu, Lin, Lin, Ma, Sheng, Tong, Zhang, Zhang, Zhang, Zhang, Zhu, Zhu, Chen, Chen, Wang, Yu, Song, Wei, Zhou, Liu, Ma, Zhang, Yan, Wu, and Wang}]{yu2025dapo}
Qiying Yu, Zheng Zhang, Ruofei Zhu, Yufeng Yuan, Xiaochen Zuo, YuYue, Weinan Dai, Tiantian Fan, Gaohong Liu, Juncai Liu, LingJun Liu, Xin Liu, Haibin Lin, Zhiqi Lin, Bole Ma, Guangming Sheng, Yuxuan Tong, Chi Zhang, Mofan Zhang, and 17 others. 2026.
\newblock \href {https://openreview.net/forum?id=2a36EMSSTp} {{DAPO}: An open-source {LLM} reinforcement learning system at scale}.
\newblock In \emph{The Thirty-ninth Annual Conference on Neural Information Processing Systems}.

\bibitem[{Zheng et~al.(2024)Zheng, Yin, Xie, Sun, Huang, Yu, Cao, Kozyrakis, Stoica, Gonzalez, Barrett, and Sheng}]{zheng2024sglang}
Lianmin Zheng, Liangsheng Yin, Zhiqiang Xie, Chuyue Sun, Jeff Huang, Cody~Hao Yu, Shiyi Cao, Christos Kozyrakis, Ion Stoica, Joseph~E. Gonzalez, Clark Barrett, and Ying Sheng. 2024.
\newblock \href {https://openreview.net/forum?id=VqkAKQibpq} {{SGL}ang: Efficient execution of structured language model programs}.
\newblock In \emph{The Thirty-eighth Annual Conference on Neural Information Processing Systems}.

\end{thebibliography}

\clearpage
\appendix

\section{Training Details}
\label{sec:hyperparams}

Table~\ref{tab:hyperparams} summarizes the RL training configuration. We use Megatron-LM~\citep{shoeybi2019megatron} as the training backend and SGLang~\citep{zheng2024sglang} for rollout. All experiments are conducted on H100 GPUs. Training wall-clock time per experiment is $\sim$22 hours for Qwen3-4B-Thinking-2507 (100 steps, 16 GPUs), $\sim$50 hours for Qwen3-8B (100 steps, 16 GPUs), and $\sim$14 hours for Qwen3-30B-A3B-Thinking-2507 (50 steps, 32 GPUs).

During the training, we use an LLM-as-judge to score CrossEntity and LongMath (\S\ref{sec:rl}). The judge prompt is shown in Figure~\ref{fig:train_judge_prompt}.

\begin{table}[H]
\centering
\small
\begin{tabular}{lc}
\toprule
\textbf{Hyperparameter} & \textbf{Value} \\
\midrule
Optimizer & Adam \\
Learning rate (4B-Thinking, 8B) & $1\text{e-}6$ \\
Learning rate (30B-A3B-Thinking) & $2\text{e-}6$ \\
LR schedule & constant \\
Global batch size (prompts) & $128$ \\
PPO batch size (responses) & $1{,}024$ \\
Rollouts per prompt ($G$) & $8$ \\
Rollout temperature & $1.0$ \\
Rollout top-$p$ & $0.95$ \\
Max input length & $64$K tokens \\
Max generation length & $16$K tokens \\
Training steps (4B-Thinking, 8B) & $100$ \\
Training steps (30B-A3B-Thinking) & $50$ \\
\bottomrule
\end{tabular}
\caption{RL training hyperparameters.}
\label{tab:hyperparams}
\end{table}

\begin{figure}[htbp]
\centering
\includegraphics[width=\columnwidth]{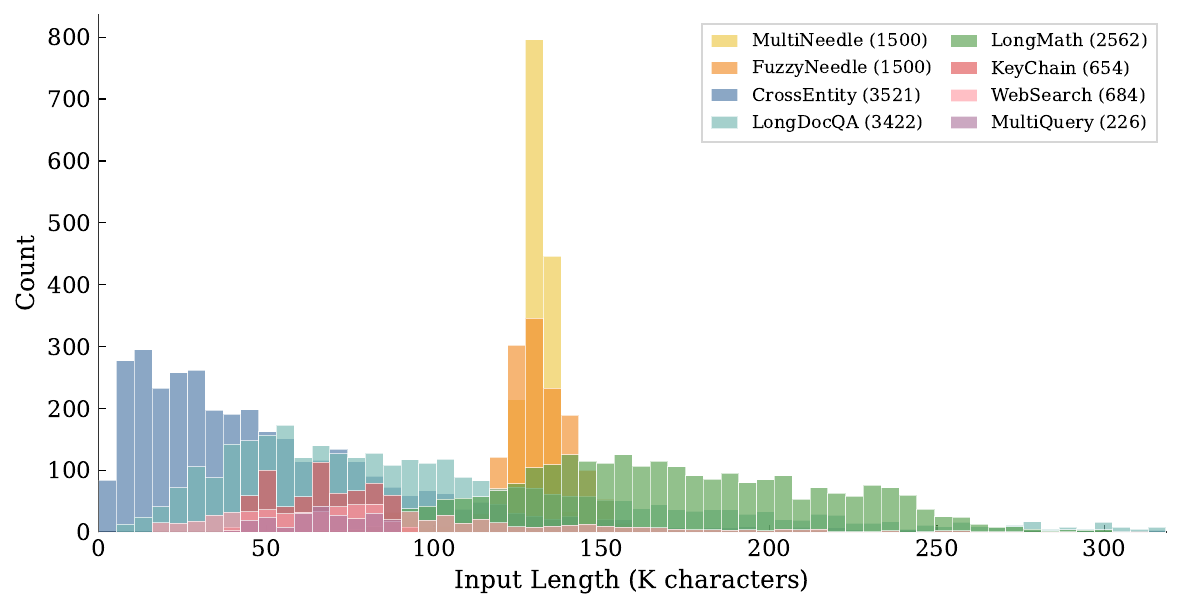}
\caption{Input length distribution (in characters) of each dataset in our training mixture.}
\label{fig:length_distribution}
\end{figure}

\paragraph{Data length distribution.}
Figure~\ref{fig:length_distribution} shows the input length distribution of each dataset in our training mixture.

\begin{figure}[htbp]
\centering
\begin{tcolorbox}[
  colback=gray!5,
  colframe=gray!60,
  fontupper=\footnotesize\ttfamily,
  title={\footnotesize\textbf{Training Judge Prompt (CrossEntity \& LongMath)}},
  boxrule=0.5pt,
  arc=2pt,
  left=4pt, right=4pt, top=4pt, bottom=4pt,
]
You are an intelligent judge who evaluates the correctness of a model's prediction against a standard answer.\par\medskip
Rules:\\
1. Ignore minor formatting differences (e.g., punctuation, case, extra whitespace).\\
2. The model's prediction may contain reasoning steps. Focus only on the final answer or conclusion.\\
3. If the prediction matches the standard answer in meaning, it is correct.\\
4. If the standard answer is short (e.g., ``Yes''), and the prediction is ``Yes, because...'', it is CORRECT.\par\medskip
Output Format:\\
If the prediction is correct, output exactly {[[1]]}.\\
If the prediction is incorrect, output exactly {[[0]]}.\\
Do not output any other text or explanation.
\end{tcolorbox}
\caption{LLM-as-judge prompt used during RL training to score CrossEntity and LongMath.}
\label{fig:train_judge_prompt}
\end{figure}

\section{Evaluation Details.}
\paragraph{Evaluation Configurations.}
We use DeepSeek-V3.2 as the LLM judge throughout all evaluations. For LongBench~v1 QA and FRAMES, we report the maximum of the LLM-judge score and substring exact match; the judge is queried with the prompt shown in Figure~\ref{fig:eval_judge_prompt}. For DocFinQA and AA-LCR, we similarly use the LLM judge to assess whether the model's prediction matches the gold answer. For LongBench~v2, LongReason, and GraphWalks, we apply rule-based answer extraction and comparison.

\begin{figure}[htbp]
\centering
\begin{tcolorbox}[
  colback=gray!5,
  colframe=gray!60,
  fontupper=\footnotesize\ttfamily,
  title={\footnotesize\textbf{Eval Judge Prompt: LongBench v1 QA \& FRAMES}},
  boxrule=0.5pt,
  arc=2pt,
  left=4pt, right=4pt, top=4pt, bottom=4pt,
]
You are a helpful assistant that judges the correctness of a model's prediction against a standard answer.\par\medskip
Question: \{question\}\\
Standard Answer: \{answer\}\\
Model Prediction: \{pred\}\par\medskip
Please judge the correctness of the model prediction and output 1 if the answer is correct, 0 if the answer is incorrect. Be fair and objective.\par\medskip
Attention:\\
1. The Model Prediction may contain a reasoning process (e.g., within <think> tags). You should focus on the final answer.\\
2. If the reasoning process is incomplete (e.g., cut off) and no final answer is provided, judge it as incorrect (0).\\
3. If a final answer is provided and matches the standard answer (even if the reasoning is cut off or lengthy), judge it as correct (1).\par\medskip
Output Format:\\
<score>1 or 0</score>
\end{tcolorbox}
\caption{LLM-as-judge prompt used for  evaluation.}
\label{fig:eval_judge_prompt}
\end{figure}

\begin{figure}[t]
\centering
\begin{tcolorbox}[
  colback=gray!5,
  colframe=gray!60,
  fontupper=\footnotesize\ttfamily,
  title={\footnotesize\textbf{Eval Judge Prompt: AA-LCR \& DocFinQA}},
  boxrule=0.5pt,
  arc=2pt,
  left=4pt, right=4pt, top=4pt, bottom=4pt,
]
Please act as an impartial judge and evaluate whether the predicted answer is consistent with the gold standard answer.\par\medskip
<Question>\\
\{question\}\\
</Question>\par\medskip
<Gold Answer>\\
\{gold\}\\
</Gold Answer>\par\medskip
<Predicted Answer>\\
\{model\_prediction\}\\
</Predicted Answer>\par\medskip
Evaluation criteria:\\
1. The gold answer is definitely correct. Do NOT re-answer the question.\\
2. For numerical answers, values that are numerically equal or very close are considered correct (e.g., 0.16 vs 0.16000, 34.8\% vs 34.81\%, \$1.2M vs 1,200,000, -3.22\% vs -3.2\%).\\
3. Different but equivalent representations are acceptable (e.g., 94 vs 94.0, 14\% vs 0.14).\\
4. For multi-part answers, all parts must match.\\
5. If no clear answer can be extracted from the prediction, grade as INCORRECT.\par\medskip
Output ONLY your verdict:\\
{[[A]]} if the predicted answer is CORRECT\\
{[[B]]} if the predicted answer is INCORRECT
\end{tcolorbox}
\caption{LLM-as-judge prompt used for AA-LCR and DocFinQA evaluation.}
\label{fig:eval_judge_prompt_b}
\end{figure}

\paragraph{Detailed scores on HELMET.}
We report detailed HELMET scores at different input lengths in Table~\ref{tab:helmet_details}.

\begin{table*}[!t]
\centering
\small
\setlength{\tabcolsep}{5pt}
\renewcommand{\arraystretch}{1.1}
\resizebox{\textwidth}{!}{%
\begin{tabular}{l c c c c c c c c c}
\toprule
\textbf{Model} & \textbf{Length} & \textbf{Recall} & \textbf{RAG} & \textbf{ICL} & \textbf{Cite} & \textbf{Re-rank} & \textbf{LongQA} & \textbf{Summ.} & \textbf{Avg.} \\
\midrule
\multirow{3}{*}{Qwen3-4B-Thinking-2507} & 32k  & 99.94 & 67.00 & 51.96 & 35.03 & 69.09 & 41.15 & 30.99 & 56.45 \\
 & 64k  & 99.81 & 69.21 & 41.20 & 25.11 & 49.10 & 42.90 & 33.54 & 51.55 \\
 & 128k & 90.00 & 62.96 & 10.44 & 7.72 & 36.95 & 49.08 & 34.31 & 41.64 \\
\midrule
\multirow{3}{*}{\quad + Ours} & 32k  & 99.75 & 72.59 & 50.20 & 33.32 & 71.03 & 44.38 & 32.32 & 57.66 \\
 & 64k  & 99.25 & 75.25 & 45.16 & 22.24 & 55.97 & 42.82 & 34.00 & 53.53 \\
 & 128k & 87.19 & 73.08 & 15.12 & 7.67 & 37.34 & 49.54 & 36.45 & 43.77 \\
\bottomrule
\end{tabular}%
}
\caption{Detailed HELMET scores of Qwen3-4B-Thinking-2507 and our RL-trained variant at different input lengths.}
\label{tab:helmet_details}
\end{table*}

\begin{table*}
\centering
\small
\setlength{\tabcolsep}{5pt}
\renewcommand{\arraystretch}{1.15}
\begin{tabular}{c c c c c c c c c}
\toprule
\textbf{Model} & \textbf{Avg.} & \textbf{LBv1-QA} & \textbf{FRAMES} & \textbf{LBv2} & \textbf{AA-LCR} & \textbf{DocFinQA} & \textbf{LongReason} & \textbf{GraphWalks} \\
\midrule
AgentCPM-Explore         & 45.57          & 62.00          & 76.33          & 29.49          & 10.25          & 39.91          & 60.08          & 40.91 \\
\midrule
\quad + Ours (25 steps)  & 52.92          & 68.50          & 75.00          & 38.17          & 31.00          & 44.03          & 65.62          & \textbf{48.13} \\
\quad + Ours (50 steps)  & \textbf{54.64} & \textbf{69.30} & \textbf{75.85} & \textbf{41.35} & \textbf{31.50} & \textbf{47.61} & \textbf{70.28} & 46.58 \\
\bottomrule
\end{tabular}
\caption{Long-context benchmark results for AgentCPM-Explore and its RL-trained variants. \textbf{Bold} denotes the best score per column among the RL-trained variants.}
\label{tab:agent_longctx}
\end{table*}

\section{Agent Evaluation Details}
\label{sec:agent_eval}

\subsection{Agent Evaluation Configurations}

We adopt the same evaluation framework (AgentToLeaP) as AgentCPM-Explore~\citep{AgentCPMExplore2026}.
The agent is equipped with three tools: (1)~a web search tool backed by Google Serp API, (2)~a URL fetching tool with a DeepSeek-V3.2 browser processor that summarizes retrieved web content, and (3)~a sandboxed Python code executor. The inference engine (SGLang) is configured with a 240k-token context window. The maximum output tokens per interaction is set to 16{,}384 for both GAIA and BrowseComp, with a maximum of 200 interaction rounds per task.

For GAIA, we evaluate on the text-only subset of the validation set, totaling 127 examples. For BrowseComp, we evaluate on a 100-example subset sampled from the full test set due to the time cost of multi-round agentic interactions.

\paragraph{Leakage filtering.}
It is a known issue that GAIA test cases are frequently leaked online~\citep{han2025stc}, often appearing in blog posts or papers that use GAIA as an illustrative example. To prevent search tools from directly retrieving leaked answers during evaluation, we implement a real-time leakage filter: when a search or fetch results simultaneously contains a substring matching the original benchmark question and the gold answer verbatim, we filter it out and do not provide it to the agent. 

\subsection{Additional Benchmark Details}

\begin{table}[!ht]
\centering
\small
\setlength{\tabcolsep}{4pt}
\renewcommand{\arraystretch}{1.15}
\begin{tabular}{l cc}
\toprule
\textbf{Model} & \textbf{GAIA} & \textbf{BrowseComp} \\
\midrule
AgentCPM-Explore         & 49.6          & 17.3 \\
\quad + Ours (25 steps)  & 50.7          & 17.8 \\
\quad + Ours (50 steps)  & \textbf{54.9} & \textbf{19.7} \\
\bottomrule
\end{tabular}
\caption{Average Pass@3 on GAIA and BrowseComp.}
\label{tab:agent_no_cm}
\end{table}

\paragraph{Per-benchmark long-context scores.}
We report the per-benchmark long-context scores of AgentCPM-Explore and its RL-trained variants in Table~\ref{tab:agent_longctx}, complementing the averaged \textbf{Avg.\ LC} column of Table~\ref{tab:agent}.

\begin{table}[!ht]
\centering
\small
\setlength{\tabcolsep}{6pt}
\renewcommand{\arraystretch}{1.15}
\resizebox{\linewidth}{!}{%
\begin{tabular}{c c c c}
\toprule
\textbf{Model} & \textbf{L1 (n=42)} & \textbf{L2 (n=66)} & \textbf{L3 (n=19)} \\
\midrule
AgentCPM-Explore         & 78.6 & 65.2 & 42.1 \\
\quad + Ours (25 steps)  & 81.0 & 63.6 & \textbf{52.6} \\
\quad + Ours (50 steps)  & \textbf{85.7} & \textbf{66.7} & \textbf{52.6} \\
\bottomrule
\end{tabular}%
}
\caption{GAIA Pass@3 broken down by official difficulty level.}
\label{tab:agent_gaia_level}
\end{table}

\begin{table}[!ht]
\centering
\small
\setlength{\tabcolsep}{4pt}
\renewcommand{\arraystretch}{1.15}
\resizebox{\linewidth}{!}{%
\begin{tabular}{c c c c c}
\toprule
\textbf{Model} & \textbf{Easy} & \textbf{Med} & \textbf{Hard} & \textbf{VHard} \\
& \textit{(n=14)} & \textit{(n=9)} & \textit{(n=21)} & \textit{(n=56)} \\
\midrule
AgentCPM-Explore         & 100.0 & 77.8 & 28.6 & 0.0 \\
\quad + Ours (25 steps)  & 100.0 & 77.8 & 38.1 & 0.0 \\
\quad + Ours (50 steps)  & 100.0 & \textbf{88.9} & \textbf{57.1} & 0.0 \\
\bottomrule
\end{tabular}%
}
\caption{BrowseComp Pass@3 by consensus difficulty. Difficulty is determined by the number of correct answers across 3 models $\times$ 3 rounds = 9 trials: easy (6--9), med (3--5), hard (1--2), vhard (0).}
\label{tab:agent_browsecomp_level}
\end{table}

\paragraph{Performance by question difficulty.}
We further stratify Pass@3 by question difficulty. For GAIA we use the official L1--L3 levels; for BrowseComp, which provides no official labels, we apply a consensus-based bucketing across $3$ models $\times$ $3$ rounds $= 9$ trials per question, labelling questions as \textit{easy} ($6$--$9$ correct), \textit{med} ($3$--$5$), \textit{hard} ($1$--$2$), and \textit{vhard} ($0$). Results are reported in Tables~\ref{tab:agent_gaia_level} and~\ref{tab:agent_browsecomp_level}. The gains of our $50$-step checkpoint are most pronounced on the harder subsets---$42.1 \to 52.6$ on GAIA L3 and $28.6 \to 57.1$ on BrowseComp Hard.

\section{Dataset Licenses}
Table~\ref{tab:licenses} lists the licenses of all external datasets used in this work. Our training recipe uses only the official training splits of these datasets. Remaining datasets in our mixture  are constructed by us and will be released under a permissive license. The LoongRL KeyChain training data~\citep{wang2026loongrl} does not specify an explicit license in its repository, though its underlying sources are all under permissive licenses.

\begin{table}[htbp]
\centering
\small
\setlength{\tabcolsep}{4pt}
\renewcommand{\arraystretch}{1.15}
\begin{tabular}{@{}llll@{}}
\toprule
\textbf{Dataset} & \textbf{Usage} & \textbf{License} \\
\midrule
HotpotQA          & LongDocQA & CC-BY-SA-4.0 \\
MuSiQue           & LongDocQA & CC-BY-4.0 \\
Qasper            & LongDocQA & CC-BY-4.0 \\
DeepMath-103K     & LongMath  & MIT \\
MATH-Hard         & LongMath  & MIT \\
\midrule
DocQA-RL-1.6K     & Baseline         & Apache-2.0 \\
LoongRL KeyChain  & Baseline         & Not specified \\
\bottomrule
\end{tabular}
\caption{Licenses of external datasets.}
\label{tab:licenses}
\end{table}

\end{document}